\documentclass[graybox]{svmult}

\usepackage{type1cm}        
\usepackage{makeidx}         
\usepackage{graphicx}        
\usepackage{multicol}        
\usepackage[bottom]{footmisc}

\usepackage{newtxtext}       %
\usepackage[varvw]{newtxmath}       

\makeindex             

\usepackage{color}

\usepackage{subfig}

\newcommand{\EE}{\mathbb{E}}
\newcommand{\VV}{\mathbb{V}}
\newcommand{\PP}{\mathbb{P}}
\newcommand{\Cov}{\mathbb{C}\mathrm{ov}}
\newcommand{\R}{\mathbb{R}}
\newcommand{\N}{\mathbb{N}}

\DeclareMathOperator*{\argmin}{arg\,min}

\newcommand{\dhh}{d_{\mbox {\tiny{\rm Hell}}}}

\begin{document}

\title*{Introduction To Gaussian Process Regression In Bayesian Inverse Problems, With New Results On Experimental Design For Weighted Error Measures}
\titlerunning{GP Regression in Inverse Problems}
\author{T. Helin, A.M. Stuart, A.L. Teckentrup, K.C. Zygalakis}
\institute{Tapio Helin \at LUT University, School of Engineering, Yliopistokatu 4, P.O. Box 20, 
FI-53851 Lappeenranta, Finland , \email{tapio.helin@lut.fi}
\and Andrew M Stuart \at California Institute of Technology,  Computing and Mathematical Sciences, 1200 E. California Blvd., MC 305-16, Pasadena, CA 91125, USA \email{astuart@caltech.edu}
\and Aretha L Teckentrup \at University of Edinburgh, School of Mathematics, James Clerk Maxwell Building, Peter Guthrie Tait road, EH9 3FD, Scotland \email{a.teckentrup@ed.ac.uk}
\and Konstantinos C Zygalakis \at University of Edinburgh, School of Mathemats, James Clerk Maxwell Building, Peter Guthrie Tait road, EH9 3FD, Scotland \email{k.zygalakis@ed.ac.uk}}
%
%
\maketitle

\abstract*{Bayesian posterior distributions arising in modern applications, including inverse problems in partial differential equation models in tomography and subsurface flow, are often computationally intractable due to the large computational cost of evaluating the data likelihood. To alleviate this problem, we consider using Gaussian process regression to build a surrogate model for the likelihood, resulting in an approximate posterior distribution that is amenable to computations in practice. This work serves as an introduction to Gaussian process regression, in particular in the context of building surrogate models for inverse problems, and presents new insights into a suitable choice of training points. We show that the error between the true and approximate posterior distribution can be bounded by the error between the true and approximate likelihood, measured in the $L^2$-norm weighted by the true posterior, and that efficiently bounding the error between the true and approximate likelihood in this norm suggests choosing the training points in the Gaussian process surrogate model based on the true posterior.}

\abstract{
Bayesian posterior distributions arising in modern applications
are often computationally intractable due to the large computational 
cost of evaluating the data likelihood. Examples include inverse problems in partial differential equation models arising in climate modeling and in subsurface fluid flow. {To alleviate the problem of expensive likelihood evaluation,} a natural
approach is to use Gaussian process regression to build a surrogate model for the likelihood, resulting in an approximate posterior distribution that is amenable to computations in practice. This paper serves as an introduction to Gaussian process regression, in particular in the context of building surrogate models for inverse problems; we also present new insights into a suitable choice of training points,
motivated by the use of Gaussian processes in approximate Bayesian inversion.
We show that the error between the true and approximate posterior distribution can be bounded by the error between the true and approximate likelihood, measured in the $L^2$-norm weighted by the true posterior; furthermore we show that minimizing the error between the true and approximate likelihood in this norm suggests choosing the training points in the Gaussian process surrogate model based on the true posterior.}

\section{Introduction}
Parameters in mathematical models in science and engineering are often not fully known and have to be {estimated from observed data}. Accurate reconstruction of the parameters, as well as an estimate of the uncertainty in the reconstruction, are crucial for reliable predictions and risk assessments. The recent explosion of available data, driven by the increase in large-scale scientific experiments and the development of sensor technology, means that there is a pressing need to develop new algorithms for the seamless integration of observed data with sophisticated mathematical models.

The complexity of models involved in modern applications, such as those based on partial differential equations, result in a {huge computational cost} and render many methods for solving inverse problems practically infeasible. A widespread solution to this problem is to approximate the model by a computationally cheaper surrogate model to obtain an approximate likelihood that is much faster to {compute, feasible to use for simulations in practice and ideally is accurate where it needs to be accurate for solution of the parameter learning problem of interest} \cite{kennedy2001bayesian,mnr07,bwg08,mx09,cfo11,st18}.

This paper discusses the use of surrogate models in the Bayesian approach to inverse problems, in which we find the {\em posterior} distribution of the unknown parameters conditioned on the observed data. A particular focus is on random surrogate models and Gaussian process regression. We provide new results that show that the context of the surrogate model, i.e. the Bayesian posterior distribution, should be taken into account when designing the surrogate model, by choosing training points in the area of parameter space where the posterior places significant mass. This poses a somewhat circular problem, since the reason we are using Gaussian process regression is to be able to compute the posterior. However, various computational approaches have been suggested to circumvent this problem in practice, including  a sequential design strategy that only requires access to the approximate posterior \cite{sn17}, and the updating of training points while exploring the posterior with sampling methods such as Markov chain Monte Carlo \cite{dunbar2021calibration,cleary2021calibrate}.


The structure of the remainder of the paper is the following. In section \ref{sec:inv}, we introduce Bayesian posterior distributions in inverse problems, discuss computational approximations using surrogate models, and prove new error bounds between the true and approximate posterior distributions in a norm weighted by the posterior. The main novel results are Theorems \ref{thm:mean_posterior} and \ref{thm:marginal_posterior}. In section \ref{sec:gp}, we introduce Gaussian process regression as surrogate models, and prove new results on the accuracy of Gaussian process regression and suitable choices of training points in the context of inverse problems. The main new results are Corollary \ref{cor:post_conv_gp_phi}, Corollary \ref{cor:post_conv_gp_G} and Theorem \ref{thm:gp_conv_weighted}. Section \ref{sec:num} gives some simple numerical examples illustrating the theoretical findings.

\section{Bayesian Inverse Problems and Their Approximation}\label{sec:inv}
{In this section, we set up the Bayesian inverse problem and describe its approximation using random surrogate models. In subsection \ref{ssec:BIP} we introduce
the framework for Bayesian inversion, and discuss MCMC methods for them.
Subsection \ref{ssec:exbip} contains two large-scale examples that motivate 
the need for surrogate modeling. In subsection \ref{ssec:inv_surrogate} we
explain surrogate modeling in detail leading, in  subsection \ref{ssec:justify},
to error estimates summarizing the effect of errors in the surrogate model on
the solution of the Bayesian inverse problem.}

\subsection{Bayesian Inverse Problems}
\label{ssec:BIP}
We are interested in solving the inverse problem of determining an unknown parameter $u \in U$ from noisy, indirect data $y \in \R^{d_y}$ given by
\begin{equation}
\label{eq:naa}
y = \mathcal G(u) + \eta,
\end{equation}
for some observation operator $\mathcal G : U \rightarrow \R^{d_y}$. For ease of presentation we assume the noise $\eta$ is a realization of the $\mathbb R^{d_y}$-valued Gaussian random variable $\mathcal N(0,\Gamma)$, for a known positive definite covariance matrix $\Gamma$, but other distributions on $\eta$ can be dealt with similarly \cite{lst18}.

{We adopt a Bayesian statistical perspective, in which  the pair $(u,y)$ is treated as a random variable ($u$ in finite dimensions) or a random process ($u$ in infinite dimensions).} The aim is to find the distribution of the conditional random variable $u|y$. This approach leads to a well-posed problem in the sense of Hadamard: there exists a unique conditional distribution $u|y$ that depends continuously on $y$ \cite{stuart10, latz20}.

In the absence of data, we assume $u$ is distributed according to a prior measure $\mu_0$. {Equation \eqref{eq:naa} then defines the conditional distribution of $y|u$, assuming that $\eta$ is independent of $u.$ Suitable choices of prior measure will depend on the application. We are then interested in the posterior distribution $\mu^y$ on the conditioned random variable $u | y$, which can be characterized as follows through  Bayes' Theorem. This delivers the
the Radon-Nikodym derivative of the posterior
with respect to the prior distribution (see e.g. \cite{stuart10}).}

\begin{proposition} \label{prop:bayes} Suppose $U$ is a separable Banach space, $\mathcal G : U \rightarrow \R^{d_y}$ is continuous and $\mu_0(U) = 1$. Then the posterior distribution $\mu^y$ on the conditioned random variable $u | y$ is absolutely continuous with respect to $\mu_0$ and given by Bayes' Theorem: 
\begin{equation*}
\frac{d\mu^y}{d\mu_0}(u) = \frac{1}{Z} \exp\big(-\Phi(u)\big),
\end{equation*}
where
\begin{align*}
\Phi(u) = \frac{1}{2} \left\| y -  \mathcal G (u) \right\|_{{\Gamma}}^2, \qquad
\text{and } \qquad 
Z = \EE_{\mu_0}\Big(\exp\big(-\Phi(u)\big)\Big).
\end{align*}
\end{proposition}

In the preceding we adopt the notational convention
{$\|\cdot \|_{A}=\|A^{-\frac12}\cdot\|_2$}
as in \cite{stuart10}, where $\|\cdot\|_2$ is the Euclidean norm and $A$ is any symmetric positive matrix.
In a finite-dimensional setting, where $u \in U \subseteq \R^{d_u}$ and we are inferring a finite number of unknown parameters, Bayes' Theorem can be written in terms of the probability density function (pdf) of the prior and posterior, denoted by $\pi_0$ and $\pi^y$, respectively \cite{kaipio2005statistical}. This takes the form
\[
\pi^y (u) = \frac{1}{Z} \exp\big(-\Phi(u)\big) \pi_0(u),
\]
with $\Phi$ and $Z$ as defined in Proposition \ref{prop:bayes}. 

The term $\exp\big(-\Phi(u)\big)$ is referred to as the {\em data likelihood}, and comes from the distribution of $y|u$. In other words, it characterizes how likely it is to observe the data $y$ given a particular choice of the parameter $u$. Since $y = \mathcal G(u) + \eta$ and $\eta \sim \mathrm{N}(0,\Gamma)$, we have $y|u \sim \mathrm N(\mathcal G(u),\Gamma)$, and the pdf of $y|u$ is hence proportional to $\exp\big(-\Phi(u)\big)$. The normalization constant $Z$ ensures that the posterior $\mu^y$ is a probability distribution, with $\mu^y(U)=1$. By Bayes' Theorem, $Z = Z(y)$ is the marginal pdf of the data $y$, and hence characterizes how likely it is to observe the data $y$ given the observational model $\mathcal G$. $Z$ is therefore often referred to as the {\em model evidence}. The analytical value of $Z$ is usually not known, and computing $Z$ numerically is notoriously difficult (see e.g. \cite{zja16}).

In applications, the goal is usually to compute a quantity of interest related to the posterior distribution $\mu^y$. This could for example be an expected value $\EE_{\mu^y}[g(u)]$, where $g : U \rightarrow U$ is chosen as the identity for the conditional mean $\EE[u|y] = \EE_{\mu^y}[u]$ or $g: U \rightarrow \{0,1\}$ is chosen as the indicator function $\mathrm{I}_{ u \in A}$ for computing event probabilities $\PP[u|y \in A] = \EE_{\mu^y}[\mathrm{I}_{ u \in A}]$.
The method of choice for sampling from the posterior distribution, enabling the computation of expected values and other quantities of interest, is often Markov chain Monte Carlo (MCMC) \cite{hastings70,mrrtt53,robert_casella,conrad2016accelerating,gc11,crsw13}. A prototypical example is the Metropolis-Hastings algorithm, which consists of the following steps for sampling from a target density $\pi$ on a finite-dimensional parameter space $U \subseteq \R^{d_u}$:

\begin{itemize}
\item[1.] Choose $u^{(1)}$ with $\pi(u^{(1)}) > 0$. 
\item[2.] At state $u^{(i)}$, sample a proposal $u'$ from
density $q(u' \, | \, u^{(i)})$.
\item[3.] Accept sample $u'$ with probability \vspace{-1ex}
\[
\alpha(u' \, | \, u^{(i)}) = \displaystyle \min \bigg( 1,
    \frac{\pi(u') \, q(u^{(i)} \, | \, u')}{\pi(u^{(i)}) \, q(u' \, | \, u^{(i)})}\bigg),
\] \\ \vspace{-1.5ex}
i.e. $u^{(i+1)}=u'$ with probability
$\alpha(u' \, | \, u^{(i)})$; otherwise stay at $u^{(i+1)}=u^{(i)}$. 
\end{itemize}

Steps 2 and 3 are repeated until the required number of samples have been generated. In the context of inverse problems and Bayesian posterior distributions, we note in particular that knowledge of the normalization constant of the target density $\pi$ is not required, since this cancels in the ratio in $\alpha$. The crucial ingredient in the algorithm is the {\em proposal density} $q$ in step 2, and a wide range of options exists, from simple random walks to methods using (higher-order) derivatives of the target and the geometry of the parameter space (see e.g. \cite{robert_casella,crsw13,gc11}). There is a rich theory underpinning the Metropolis-Hastings algorithm, and in particular, it is guaranteed under mild assumptions that the distribution of $u^{(i)}$ converges to the target density $\pi$ as $i \rightarrow \infty$ (see e.g. \cite{robert_casella}).

MCMC methods typically require repeated evaluation of the data likelihood. In the Metropolis-Hastings algorithm above; this can be seen in step 3, where $\exp\big( - \Phi(u') \big)$ needs to be computed for every proposal $u'$. This quickly becomes infeasible in modern applications where the computation of the likelihood is very costly. This includes for example inverse problems in geophysics (subsurface flow model) and climate (general circulation model), where $\mathcal G$ involves the solution of one or more coupled partial differential equations (PDEs). Two such examples are given 
in the next subsection.

\subsection{Examples of Large-Scale Complex Bayesian Inverse Problems}
\label{ssec:exbip}

\begin{example} {\em Subsurface Flow Model} {A simple model for stationary subsurface fluid flow is given by
\begin{equation} \label{eq:pde}
- \nabla \cdot \left( k(x) \nabla p(x) \right) = g(x), \qquad x \in D,
\end{equation}
where $k$ represents the permeability (or hydraulic conductivity) of the subsurface, $p$ denotes the pressure head of the fluid, $g$ incorporates any sources or sinks, and suitable boundary conditions are imposed on the computational domain $D$ \cite{zhang,rubin}. This model comes from a combination of Darcy's law for single-phase fluid flow in a porous medium, $q(x) = - k(x) \nabla p(x)$, and conservation of mass, $ \nabla \cdot q = g$. 

A typical inverse problem in this context is to reconstruct the permeability $k$ given noisy measurements of the pressure head $y = \{ p(x_i) + \eta_i\}_{i=1}^{d_y}$ or the Darcy flux $y = \{ k(x_i) \nabla p(x_i) + \eta_i\}_{i=1}^{d_y}$ at fixed locations $x_i \in D$. Although we in general wish to reconstruct the function $k \in L^2(D)$, we often choose a parametrization of $k$ in computations. The inverse problem then becomes that
of learning the coefficients $u \in \R^{d_u}$ in the parametrization. For example, we can choose a piecewise constant model
\[
k(x; u) = \sum_{j=1}^{d_u} {u_j} \mathrm{I}_{x \in D_j}(x),
\]
where $D_j$ is a partitioning of the domain $D$ representing layers of different types of rock, and $u_j$ is the value of the permeability $k$ in the layer $D_j$. Since the permeability is always positive, and can vary over orders of magnitude between different types of rock, the prior distribution $\mu_0$ on $u =\{u_j\}_{j=1}^{d_u}$ should reflect these properties, and log-normal distributions are often used. 

Evaluation of the parameter-to-observation map $\mathcal G : \R^{d_u}_{> 0} \rightarrow \R^{d_y}$, defined by $\mathcal G(u) = \{ p(x_i; u) \}_{i=1}^{d_y}$ or $\mathcal G(u) = \{ k(x_i; u) \nabla p(x_i; u) \}_{i=1}^{d_y}$, requires the solution of the PDE \eqref{eq:pde} with a different coefficient $k$ for every $u$ and is thus typically very costly.
}
\end{example}

\begin{example} {\em General Circulation Model}
General Circulation Models (GCMs) typically comprise discretization of a system of three space-dimensional time-dependent PDEs, with
spatial dimensions representing Earth's longitude, latitude and vertical dimensions. Unknown
parameters appear in sub-grid scale models which attempt to capture unresolvable physics
on the scale of clouds. These parameters can in principle be learned from data in the form
of time-averaged satellite measurements of the Earth's atmosphere. 
Uncertainty quantification, and hence the Bayesian approach, is crucial as estimation (and reduction) of uncertainty in climate models is a central goal of modern climate modeling research \cite{schneider2020earth}. In this paper we will consider a specific GCM from \cite{o2008hydrological}.
The conservation laws, which are discretized to form the computational model, include two
equations which take the form
\begin{align*}
    \text{Moisture Conservation:}\quad \frac{\partial q}{\partial t} + {v}\cdot\nabla q &= - \frac{q - q_{\textrm{ref}}(T; {\theta})}{\tau_q(q, T; {\theta})} \\
    \text{Energy Conservation:}\quad \frac{\partial T}{\partial t} + {v}\cdot\nabla T &=  \frac{T - T_{\textrm{ref}}(q, T; {\theta})}{\tau_T(q, T; {\theta})} + \cdots\,,
\end{align*}
coupled to further conservation laws for mass and momentum. The model includes two unknown
parameters: ${\theta_{RH}}$, the reference relative humidity; and ${\theta_{\tau}}$ a relaxation timescale; the functional forms of $q_{\textrm{ref}}$, $T_{\textrm{ref}},$ $\tau_q$ and $\tau_T$ 
are known. The unknown parameter is $\theta=({\theta_{RH}},{\theta_{\tau}}).$ 
The inverse problem is to learn about $\theta$, and 
uncertainty in $\theta$, from climate statistics; the paper \cite{dunbar2021calibration}
employs 30-day averages of the free-tropospheric relative humidity, of the precipitation rate, and of a measure of the frequency of extreme precipitation. The parameter-to-data map thus requires
simulation of the discretized conservation laws. This leads to ${\mathcal O}(10^2)$ noisy and indirect observations from which to learn about the $2$ unknowns. Similar, related inverse
problems may be found in \cite{huang2022iterated}[subsection 5.10], 
in which the number of unknown parameters is $4.$
\end{example}


\subsection{Surrogate Models}\label{ssec:inv_surrogate}
To make computations feasible in practice, it is common to approximate the log-likelihood $\Phi$, or the forward model $\mathcal{G}$, by a surrogate model, also known as an emulator, meta-model or reduced order model. 
A wide range of suitable surrogate models exist, see e.g. \cite{st18,lst18} and the references therein. Recently, a particular focus has been on the use of {\em random} surrogate models. These have been shown to lead to better approximations of the posterior distribution when the error in the surrogate model is large compared to the noise in the observations \cite{cgssz17,btz23}, and to improve the statistical efficiency of MCMC methods \cite{cfo11}. Examples of random surrogate models are methods from the field of probabilistic numerics \cite{cgssz17,lss19}, randomized projection methods \cite{lmbn17,njls09}, and Gaussian process emulators \cite{kennedy2001bayesian,st18}.

Given a random surrogate model for the log-likelihood $\Phi$ or the forward model $\mathcal{G}$, there are then different ways of using this to approximate the posterior distribution $\mu^y$.  We will henceforth fix our notation to using a surrogate model for $\Phi$, since using a surrogate model for $\mathcal G$ can be translated into a surrogate model for $\Phi$ using the specific form of the likelihood.
Let us denote the random surrogate model by $\Phi_N$, where $N$ denotes a discretization parameter such as the number of training points in a Gaussian process emulator or the number of grid points used in a probabilistic PDE solver. 
Using the mean $\EE[\Phi_N]$ of the surrogate model to approximate $\Phi$, we obtain the {\em mean-based approximation} \vspace{-2ex}

\begin{align}
\frac{d\mu^{y,N}_\mathrm{mean}}{d\mu_0}(u) &= \frac{1}{Z_{N}^\mathrm{mean}} \exp\big(- \EE[\Phi_N(u)] \big), \label{eq:rad_nik_mean} \\
Z_{N}^\mathrm{mean} &= \EE_{\mu_0}\Big(\exp\big(-\EE[\Phi_N(u)] \big)\Big) \nonumber.
\end{align}

Alternatively, we can use the random surrogate $\Phi_N$ directly to approximate $\Phi$, and take the expected value of the resulting random approximation of the likelihood. This gives the {\em marginal approximation} \vspace{-2ex}

\begin{align}
\frac{d\mu^{y,N}_\mathrm{marginal}}{d\mu_0}(u) &= \frac{1}{Z_{N}^\mathrm{marginal}} \EE \Big(\exp\big(-\Phi_N (u) \big)\Big), \label{eq:rad_nik_marginal} \\
Z_{N}^\mathrm{marginal} &= \EE_{\mu_0} \left(\EE\Big(\exp\big(- \Phi_N(u)\big)\Big) \right). \nonumber
\end{align}

Intuitively, the marginal approximation introduces additional uncertainty in the approximate posterior distribution, coming from the randomness of the surrogate model. One interpretation of the randomness of our surrogate model is that since we have used a finite amount of information/resources in the construction, there is still some uncertainty (or {\em error}) associated to it. Including this uncertainty in the approximate posterior distribution then allows us to take into account the error in the surrogate model when inferring the parameter $u$. In practical applications, where the accuracy of the surrogate model might be limited due to computational resources, the uncertainty (or error) in the surrogate model can be large (or comparable) to the uncertainty present in the observations $y$, and it is crucial to take this into account to avoid over-confident and biased predictions \cite{cgssz17,btz23}.

In special cases, it can be shown analytically that the marginal approximation results in a form of variance inflation in the likelihood. One such instance is when the forward model $\mathcal G$ is approximated by a Gaussian process $\mathcal G_N \sim \mathrm{GP}(m_N^\mathcal G, K_N)$ with mean $m_N^\mathcal G$ and covariance kernel $K_N$ (see section \ref{sec:gp} for more details). The mean-based approximation assumes the observational model $y = m_N^\mathcal G(u) + \eta$, whereas the marginal approximation uses 
\[
y = \mathcal G_N(u) + \eta = m_N^\mathcal G(u) + \xi(u) + \eta,
\]
with $\xi(u) \sim \mathrm{N}(0, K_N(u,u))$. The resulting approximate likelihoods are proportional to
\[
\frac{1}{\sqrt{\det{(\Gamma)}}} \exp\left(-\frac{||y - m_N^\mathcal{G}(u)||^2_{\Gamma}}{2}\right)
\]
for the mean-based approximation, and 
\[
\frac{1}{\sqrt{\det{(\Gamma + K_N(u,u) )}}} \exp\left(-\frac{||y - m_N^\mathcal{G}(u)||^2_{(\Gamma +K_N(u,u))}}{2}\right)
\]
for the marginal approximation, see e.g. \cite{cdss18,cosg17,btz23}. The difference between the two likelihoods hence depends on the relationship between $\Gamma$, the uncertainty in the observations $y$, and $K_N$, the uncertainty in the surrogate model $\Phi_N$. If $K_N$ is small compared to $\Gamma$, the two approximate posteriors will be similar. 


The approximate posterior distributions can also be motivated from a decision theoretic point of view \cite{sn17,jgvm21}. Suppose $U$ is a compact subset of $\R^{d_u}$, and we approximate the negative log-likelihood $\Phi$ by a Gaussian process $\Phi_N \sim \mathrm{GP}(m_N^\Phi, k_N)$. Then the (un-normalized) marginal approximation minimizes the expected $L^2$-loss:
\[
 \EE \Big(\exp\big(-\Phi_N (u) \big)\Big) \pi_0(u) = \argmin_{f \in L^2(U)} \; \EE \int_U \left( \exp\big(-\Phi_N (u) \big) \pi_0(u) -f(u) \right)^2 \mathrm{d} u.
\]
The (un-normalized) mean-based approximation instead minimizes the expected $L^1$-loss:
\[
\exp\big(-m^\Phi_N (u) \big)\pi_0(u) = \argmin_{f \in L^1(U)} \; \EE \int_U \left| \exp\big(-\Phi_N (u) \big) \pi_0(u) -f(u) \right| \mathrm{d} u.
\]


\begin{remark}[Alternative approximate posteriors] In addition to the approximations described above, we can also consider the {\em sample-based} approximation
\begin{align*}
\frac{d\mu^{y,N}_\mathrm{sample}}{d\mu_0}(u) &= \frac{1}{Z_{N}^\mathrm{sample}} \exp\big(-\Phi_N (u) \big) , \\
Z_{N}^\mathrm{sample} &= \EE_{\mu_0} \left( \exp\big(- \Phi_N(u)\big) \right),
\end{align*}
which results in a random approximation of the posterior \cite{st18,lst18}. If there is considerable uncertainty in the random surrogate model $\Phi_N$, the posterior approximations corresponding to different samples of $\Phi_N$ may look very dissimilar. 

An alternative marginal approximation can be defined by taking the expected value of the normalized random likelihood defined above,
\begin{align*}
\frac{d\mu^{y,N}_\mathrm{marginal'}}{d\mu_0}(u) &= \EE \left( \frac{1}{Z_{N}^\mathrm{sample}} \exp\big(-\Phi_N (u) \big) \right).
\end{align*}
However, this approximation is not easily amenable to sampling methods such as (pseudo-marginal) MCMC, since the computation of $Z_{N}^\mathrm{sample}$ is intractable. $\blacksquare$
\end{remark}

\subsection{Error in the Surrogate-accelerated Posterior Distribution}
\label{ssec:justify}
To justify the use of the approximate posterior distributions $\mu^{y,N}_\mathrm{mean}$ and $\mu^{y,N}_\mathrm{marginal}$ in practice, we want to show that they converge to the true posterior $\mu^y$ as $N \rightarrow \infty$. We will measure the error in the Hellinger distance \vspace{-1ex} 
\begin{equation*}
\dhh(\mu^y, \mu^{y,N}) = \left( \frac{1}{2} \int_{U} \left(\sqrt{\frac{d \mu^y}{d \nu}} - \sqrt{\frac{d \mu^{y,N}}{d \nu}} \right)^2 d \nu \right)^{1/2},
\end{equation*}
where $\nu$ is any measure such that $\mu^y$ and $\mu^{y,N}$ are both absolutely continuous with respect to $\nu$ {(and the value of the distance is independent of the choice of $\nu.$)
In \cite{lst18,st18, t20} bounds were obtained on $\dhh(\mu^y, \mu^{y,N}_\mathrm{mean})$ and $\dhh(\mu^y, \mu^{y,N}_\mathrm{marginal})$ in terms of $\left\|\Phi - \EE[\Phi_N] \right\|_{L^{p}_{\mu_0}(U)}$ and $\left\| \EE\bigl[ \left| \Phi - \Phi_{N} \right|^{q_1} \bigr]^{1/q_1} \right\|_{L^{q_2}_{\mu_{0}}(U)}$, respectively, for $p$, $q_1$ and $q_2$ depending on the regularity of the problem. These results suggest that choosing $\Phi_N$ to be a good approximation to $\Phi$ in all regions where the prior $\mu_0$ places significant mass is sufficient for accurate surrogate modeling in the
context of Bayesian inference. However we would expect that, in fact, $\Phi_N$ only needs to be a good approximation to $\Phi$ in regions where the posterior $\mu^y$ places significant mass. This is captured in the following results, which follow
by analyzing the Hellinger distance with the choice $\nu = \mu^y$.}

\begin{theorem}[Convergence of mean-based approximation]\label{thm:mean_posterior}  Suppose that there exist scalars $C_1, C_2 \geq 0$, independent of $N$, such that for the H\"{o}lder-conjugate exponent pair $(p_1,p_1')$, we have\footnote{For $p_1=\infty$, assumption (i) requires bounding the standard $L^\infty(U)$-norm, due to the corresponding H\"older inequality $\int_U f g \mu^y(\mathrm{d}u) \leq \sup_{u \in U } |f(u)|  \int_U |g| \mu^y(\mathrm{d}u) = \|f\|_{L^\infty(U)} \|g\|_{L^1_{\mu^y}(U)}$.} 
	\begin{enumerate}
		\item[(i)]  $\left\| \exp\left((\Phi - \EE [\Phi_N])\right) \right\|_{L^{p_1}_{\mu^y}(U)} \leq C_1(p_1)$;
		\item[(ii)]  $C_2^{-1} \leq Z_{N}^\mathrm{mean} \leq C_2$.
	\end{enumerate}
	Then there exists a constant $C_{\mathrm{Thm} \ref{thm:mean_posterior}} =C_{\mathrm{Thm} \ref{thm:mean_posterior}} (C_1,C_2, Z)$, independent of $N$, such that 
	\begin{subequations}
		\begin{align*}
			\dhh(\mu^y, \mu^{y,N}_\mathrm{mean}) &\leq C_{\mathrm{Thm} \ref{thm:mean_posterior}}  \left\|\Phi - \EE [\Phi_N] \right\|_{L^{2p_1'}_{\mu^y}(U)}.
		\end{align*}
	\end{subequations}
\end{theorem}

\begin{theorem}[Convergence of marginal approximation] \label{thm:marginal_posterior} 
	Suppose that there exist scalars $C_1, C_2 \geq 0$, independent of $N$, such that, for the H\"{o}lder-conjugate exponent pairs $(p_1,p_1')$ and $(p_2,p_2')$, we have\footnotemark[1]
	\begin{enumerate}
		\item[(i)] \label{item:hell_conv_marg_b} $\left\| \EE \big[  \left(\exp\left(\Phi - \Phi_N\right)\right)^{p_2} \big]^{1/{p_2}}  \right\|_{L^{2p_1}_{\mu^y}(U)} \leq C_1(p_1,p_2)$
		\item[(ii)] \; $C_2^{-1} \leq Z_{N}^\mathrm{marginal} \leq C_2$.
	\end{enumerate}
	Then there exists a constant $C_{\mathrm{Thm} \ref{thm:marginal_posterior}}=C_{\mathrm{Thm} \ref{thm:marginal_posterior}} (C_1,C_2,Z)$, independent of $N$, such that \vspace{-0.5ex}
	\begin{subequations}
		\label{eq:hell_conv_marg_result}
		\begin{align*}
			\dhh \bigl( \mu, \mu^{y,N}_\mathrm{marginal} \bigr) &\leq C_{\mathrm{Thm} \ref{thm:marginal_posterior}}  \left\| \EE \big[ \left(\Phi - \Phi_N \right)^{p_2'} \big]^{1/p_2'} \right\|_{L^{2p_1'}_{\mu^y}(U)}.
		\end{align*}
	\end{subequations}
\end{theorem}

Note that these results hold in the general setting where the parameter space $U$ is a Banach space, and do not require $U$ to be finite-dimensional. The proofs of these results are similar to those in \cite{st18,lst18}, and can be found in the appendix. The assumptions in the above theorems are generally applicable to any random surrogate model $\Phi_N$, and in fact also to general distributions on the noise $\eta$. They need to be checked on a case-by-case basis. This is done for Gaussian noise and surrogate models based on Gaussian process regression in section \ref{sec:gp}.

Assumption $(i)$ in Theorems \ref{thm:mean_posterior} and \ref{thm:marginal_posterior} is related to integrability properties of our surrogate model $\Phi_N$. The exponent $p_1$, present in both theorems, is related to the decay rate of the tails in $u$. The exponent $p_2$, related to the decay rate of the tails of the distribution of $\Phi_N$, is only needed in Theorem \ref{thm:marginal_posterior}, since the mean-based approximation $\mu^{y,N}_\mathrm{mean}$ does not make use of the randomness in $\Phi_N$.
Assumption $(ii)$ in Theorems \ref{thm:mean_posterior} and \ref{thm:marginal_posterior} ensures that the normalization constants $Z_{N}^\mathrm{mean}$ and $Z_{N}^\mathrm{marginal}$, and hence the approximate posteriors, are well-defined.

\section{Gaussian {Process Regression}} \label{sec:gp}
We now want to use Gaussian process regression to build a surrogate model for the data likelihood. Gaussian process regression (a.k.a Gaussian process emulation, or kriging) is a way of building an approximation to a function $f$, based on a finite number of evaluations of $f$ at a chosen set of training points (a.k.a. {design points}). 

We will here consider emulation of either the parameter-to-observation map $\mathcal G: U \rightarrow \R^{d_y}$ or the negative log-likelihood $\Phi:U \rightarrow \R$. We will for simplicity focus on the emulation of scalar valued functions, and assume that an emulator of $\mathcal G$ in the case $d_y > 1$ is constructed by emulating each entry independently. However, we remark here that in practice including correlation between different entries in $\mathcal G$ can significantly improve the quality of the approximate posterior \cite{btz23}.

{In subsection \ref{ssec:su} we set-up the framework of Gaussian process
regression, followed in subsection \ref{ssec:sm} by discussion of the use
of this regression technique in the context of surrogate modeling. Subsection
\ref{ssec:exsm} returns to the two examples from subsection \ref{ssec:exbip},
in order to illustrate the potential for reduction in computational cost achieved by the surrogate model $f_N$ compared to the original model $f$. Subsection \ref{ssec:error0}
is devoted to estimates of the error in the approximate posterior 
in terms of the error in a GP surrogate
model used to approximate the likelihood contribution to the posterior.
In subsection \ref{ssec:gp_sk} error estimates for Gaussian process regression,
which may be used in the analysis of the previous subsection, are provided.}

\subsection{Set-up}
\label{ssec:su}

Let $U \subseteq \R^{d_u}$ be finite-dimensional, and let $f : U \rightarrow \R$ be an arbitrary function. Gaussian process regression is a Bayesian procedure, and the starting point is to put a Gaussian process prior on the function $f$. In other words, we model $f$ as a sample of the Gaussian process
\begin{equation}\label{eq:gp}
{f_0} \sim \text{GP}(m(u), k(u,u')),
\end{equation}
with $m : U \rightarrow \R$ a chosen mean function, giving $ \EE[f_0(u)] = m(u)$, and $k : U \times U \rightarrow \R$  a chosen symmetric, positive-definite covariance kernel, giving $\Cov(f_0(u), f_0(u')) = k(u,u')$. This means that for any set of points $\{ \tilde u^m\}_{m=1}^M \subseteq U$, the vector $[f_0(\tilde u^1); \dots; f_0(\tilde u^M)] \in \R^M$ follows a multivariate Gaussian distribution, with mean $[m(\tilde u^1); \dots; m(\tilde u^M)] \in \R^M$ and covariance matrix $K \in \R^{M \times M}$ with entries $K_{i,j} = k(\tilde u^i, \tilde u^j)$.

Typical choices of the mean function $m$ include the zero function and polynomials \cite{rasmussen_williams}. 
A family of covariance functions $k$ frequently used in applications are the Mat\`ern covariance functions \cite{matern}, given by
\begin{equation}\label{eq:mat_cov}
k_{\nu,\lambda,\sigma_k^2}(u,u') = \sigma_k^2 \, \frac{1}{\Gamma(\nu) 2^{\nu-1}} \left(\frac{\|u-u'\|_2}{\lambda}\right)^\nu B_\nu\left( \frac{\|u-u'\|_2}{\lambda}\right) ,
\end{equation}
where $\Gamma$ denotes the Gamma function, $B_\nu$ denotes the modified Bessel function of the second kind and $\nu, \lambda$ and $\sigma_k^2$ are positive parameters. The parameter $\lambda$ is referred to as the {\em correlation length}, and governs the length scale at which ${f_0}(u)$ and ${f_0}(u')$ are correlated. The parameter $\sigma_k^2$ is the {\em marginal variance} $\sigma_k^2 = k_{\nu,\lambda,\sigma_k^2}(u,u) = \VV[f_0(u)]$, and governs the typical magnitude of ${f_0}(u)$. Finally, the parameter $\nu$ is referred to as the {\em smoothness parameter}, and governs the regularity of sample paths of ${f_0}$ as a function of $u$. Sample paths of $f_0$ are in the Sobolev space $H^{\nu-\epsilon}(U)$ almost surely, for any $\epsilon > 0$, see e.g. \cite{khss18}.

In the limit $\nu \rightarrow \infty$, we obtain the Gaussian covariance function
\begin{equation*}\label{eq:gauss_cov}
k_{\infty,\lambda,\sigma_k^2}(u,u') = \sigma_k^2 \exp \left(-\frac{\|u-u'\|_2^2}{2 \lambda^2}\right),
\end{equation*} 
also known as the squared exponential or {radial basis function (RBF)} covariance function.
The formula for the Mat\`ern covariance function furthermore simplifies when $\nu$ is a half integer, and popular choices include
\begin{align*}
k_{\frac{1}{2},\lambda,\sigma_k^2}(u,u') &= \sigma_k^2 \exp \left(-\frac{\|u-u'\|_2}{\lambda}\right), \\
k_{\frac{3}{2},\lambda,\sigma_k^2}(u,u') &= \sigma_k^2 \left( 1 + \sqrt{3} \frac{\|u-u'\|_2}{\lambda}\right)\exp \left(-\sqrt{3} \frac{\|u-u'\|_2}{\lambda}\right).
\end{align*}
The choice $\nu=\frac{1}{2}$ gives the exponential covariance function, also known as the Laplace covariance function; {in the setting of one-dimensional input
variable $u$ it is the covariance function of the Ornstein-Uhlenbeck process.}

\begin{figure}[!ht]
  \centering
  \subfloat{\includegraphics[width=.4\textwidth]{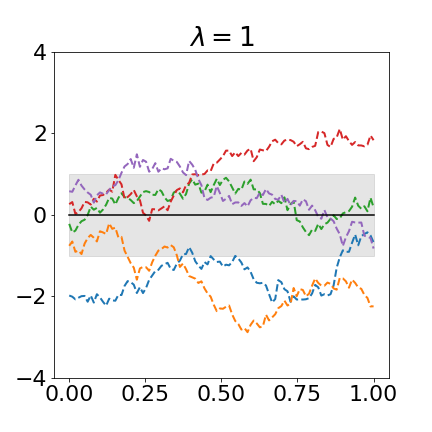}}\quad
  \subfloat{\includegraphics[width=.4\textwidth]{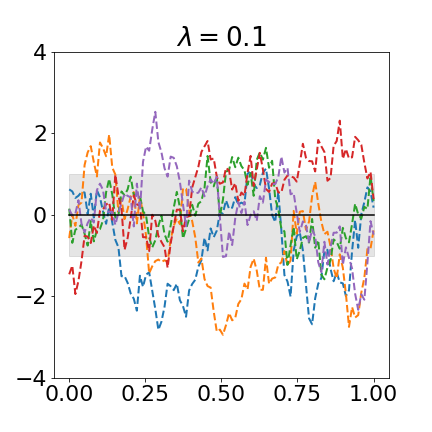}}\\
  \subfloat{\includegraphics[width=.4\textwidth]{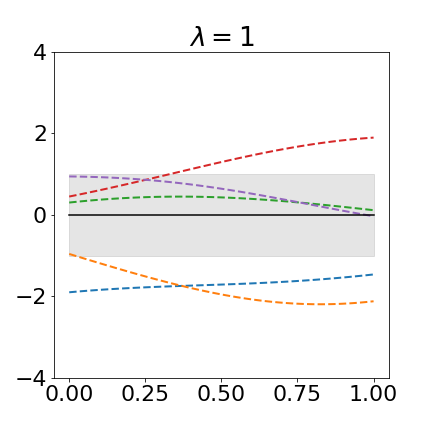}}\quad
  \subfloat{\includegraphics[width=.4\textwidth]{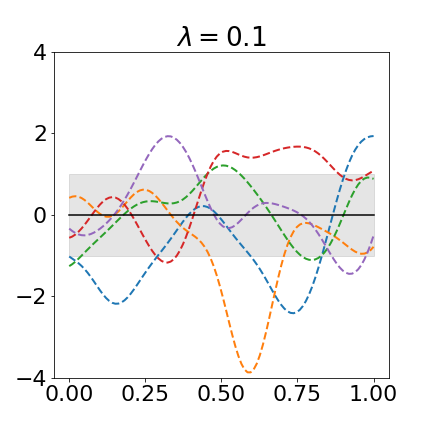}}
  \caption{Sample paths of the Gaussian process $f_0$ with Mat\'ern covariance function $k_{\nu,\lambda,\sigma_k^2}$. Top left: $\{\nu,\lambda,\sigma_k^2\} = \{\frac{1}{2},1,1\}$. Top right: $\{\nu,\lambda,\sigma_k^2\} = \{\frac{1}{2},0.1,1\}$. Bottom left: $\{\nu,\lambda,\sigma_k^2\} = \{\infty,1,1\}$. Bottom right: $\{\nu,\lambda,\sigma_k^2\} = \{\infty,0.1,1\}$.}
  \label{fig:samples}
\end{figure}

Figure \ref{fig:samples} shows 5 independent sample paths of the Gaussian process $f_0$ with $m=0$ and 4 different choices of the parameters in the Mat\'ern covariance function. We also show the mean $\EE[f(u)] = m(u) = 0$ as the solid black line, and the marginal standard deviation $\sqrt{k(u,u)} = \sigma =1$ as the light grey interval around the mean. The two plots in the top row show $\nu=\frac{1}{2}$, leading to sample paths that are continuous but not differentiable, and the two plots in the bottom row show $\nu=\infty$, in which case the sample paths are infinitely smooth. The left column shows a long correlation length $\lambda$, leading to mostly large scale variations, whereas the right column shows a shorter correlation length, resulting in more small scale fluctuations.

\begin{figure}[!ht]
  \centering
  \subfloat{\includegraphics[width=.4\textwidth]{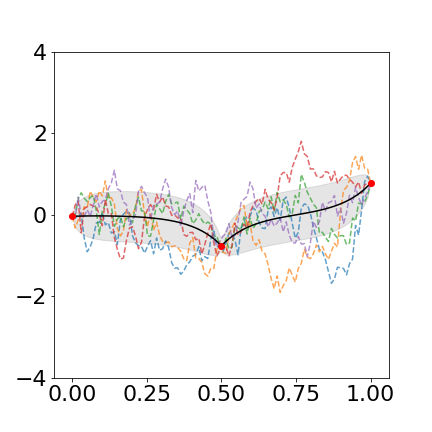}}\quad
  \subfloat{\includegraphics[width=.4\textwidth]{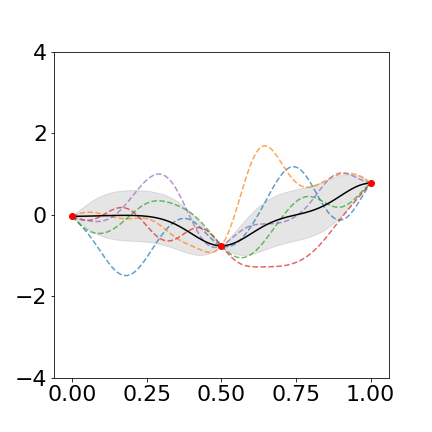}}\\
\caption{Sample paths of the Gaussian process $f_3$ with Mat\'ern covariance function $k_{\nu,\lambda,\sigma_k^2}$, for $f(x)=\sin{(x-2.5)^2}$. Left: $\{\nu,\lambda,\sigma_k^2\} = \{\frac{1}{2},0.1,0.63^2\}$ Right: $\{\nu,\lambda,\sigma_k^2\} = \{\infty,0.1,0.63^2\}$.}

  \label{fig:samples_posterior}
\end{figure}

Now suppose we are given data in the form of a set of distinct training points $D_N := \{u^n\}_{n=1}^N \subseteq U$, together with corresponding function values 
\begin{equation}\label{eq:data_exact}
f(D_N) := [f(u^1); \dots; f(u^N)] \in \R^N.
\end{equation}
{Since $f_0$ is a Gaussian process, the vector $[f_0(u^1); \dots; f_0(u^N); f_0(\tilde u^1); \dots, f_0(\tilde u^M)] \in \R^{N+M}$, for any set of test points $\{ \tilde u^m\}_{m=1}^M \subseteq U$, follows a multivariate Gaussian distribution. The conditional distribution of $f_0(\tilde u^1), \dots, f_0(\tilde u^M)$, given the values $f_0(u^1) = f(u^1), \dots, f_0(u^N) = f(u^N)$, is then again Gaussian, with mean and covariance given by the standard formulas for the conditioning of Gaussian random variables \cite{rasmussen_williams}.} 
Conditioning the Gaussian process \eqref{eq:gp} on the known values $f(D_N)$, we hence obtain another Gaussian process $f_N$, known as the {\em predictive process}. We have
\begin{equation}\label{eq:gp_pred}
f_N \sim \text{GP}(m^f_N(u), k_N(u,u')),
\end{equation}
where the predictive mean $m^f_N : U \rightarrow \R$ and predictive covariance $k_N : U \times U \rightarrow \R$ are known explicitly, and depend on the modeling choices made in \eqref{eq:gp}. We will from now on focus on the popular choice $m \equiv 0$; the case of a non-zero mean is discussed in Remark \ref{rem:mean}. 
When $m \equiv 0$, we have
\begin{align}
m_N^f(u) &= k(u,D_N)^T K(D_N,D_N)^{-1} f(D_N), \label{eq:pred_eq_mean}\\
k_N(u,u') &= k(u,u') - k(u,D_N)^T K(D_N,D_N)^{-1} k(u',D_N), \label{eq:pred_eq_var}
\end{align}
where $k(u,D_N) = [k(u,u^1); \dots; k(u,u^N)] \in \R^{N}$ and $K(D_N,D_N) \in \R^{N \times N}$ is the matrix with $ij^\mathrm{th}$ entry equal to $k(u^i,u^j)$ \cite{rasmussen_williams}. 

Figure \ref{fig:samples_posterior} is similar to Figure \ref{fig:samples}, and shows samples from the predictive process $f_N$ (with $N=3$) for different choices of hyper-parameters in the Mat\`ern covariance kernel. We also show the updated mean $\EE[f_N(u)] = m_N^f(u)$ and the updated marginal standard deviation $\sqrt{k_N(u,u)}$. We see that the choice of hyper-parameters can have a significant influence on the behaviour of the predictive process. Good values of hyper-parameters are often not known a priori, and have to be estimated from the observed values $f(D_N)$ in \eqref{eq:data_exact} (see e.g. \cite{rasmussen_williams} and the references therein). This is done for $\lambda$ and $\sigma^2$ in Figure \ref{fig:samples_posterior} using \texttt{scikit-learn} \cite{scikit-learn}.

From \eqref{eq:pred_eq_mean}, we note that $m_N^f$ interpolates the function $f$ at the training points $D_N$, since the vector $k(u^n,D_N)$ is the $n^\mathrm{th}$ row of the matrix $K(D_N,D_N)$. In other words, we have $m_N^f(u^n) = f(u^n)$, for all $n=1,\dots,N$. 
For the predictive covariance $k_N$, we note that $k_N(u,u) < k(u,u)$ for all $u \in U$, since $K(D_N,D_N)$ is positive-definite by assumption. Furthermore, we also note that $k_N(u^n,u^n) = 0$, for $n=1, \dots, N$, since $k(u^n,D_N)^T \; K(D_N,D_N)^{-1} \; k(u^n,D_N) = k(u^n,u^n)$.

\begin{remark}\label{rem:mean}{\em (Prior with non-zero mean)} If in \eqref{eq:gp} we use a non-zero mean $m(\cdot)$, the formula for the predictive mean $m_N^f$ changes to
\begin{equation*}
m_N^{f,m}(u) = m(u) + k(u,D_N)^T K(D_N,D_N)^{-1} (f(D_N) - m(D_N)),
\end{equation*}
where $m(D_N) := [m(u^1); \dots; m(u^N)] \in \R^N$. The predictive covariance $k_N(u,u')$ is as in \eqref{eq:pred_eq_mean}. As in the case $m \equiv 0$, we have $m_N^f(u^n) = f(u^n)$, for $n=1, \dots, N$, and $m_N^f$ is an interpolant of $f$. Under suitable assumptions on $m$, any error bounds derived in the case $m \equiv 0$ can be transferred to the general case, see e.g \cite{t20}. $\blacksquare$
\end{remark}

\begin{remark}\label{rem:noisy}{\em (Noisy function values)} If instead of exact function values as in \eqref{eq:data_exact}, we observe noisy function values 
\[
d := [f(u^1) + \varepsilon_1; \dots; f(u^N)+\varepsilon_N] \in \R^N,
\]
with $\varepsilon_i \sim \mathrm{N}(0,\sigma^2)$ i.i.d., then the formulas for the predictive process $f_N$ presented above hold with $K(D_N,D_N)$ replaced by $K(D_N,D_N) + \sigma^2 \mathrm{I}$. $\blacksquare$
\end{remark}

\subsection{Gaussian Process Regression as Surrogate Model}
\label{ssec:sm}
There are two main use cases for Gaussian process regression: (i) the true function $f$ generating the input-output pairs $\{u^n, f(u^n)\}_{n=1}^N$ is unknown, or (ii) the function $f$ generating $\{u^n, f(u^n)\}_{n=1}^N$ is known, but computationally very expensive to evaluate. In scenario (i), we wish to learn what the function $f$ is from the observed input-output pairs. This is often based on real data, and so the assumption that the observed function values contain noise is common. In the case of (ii), we wish to construct a surrogate model (a.k.a. reduced model, meta-model or emulator) that is cheaper to evaluate than the original $f$. This is often based on data $\{u^n, f(u^n)\}_{n=1}^N$ obtained from simulating a mathematical model on a computer, and so assuming no noise in the function values may be more {appropriate.}

To construct a surrogate model for $f$, we simulate $N$ model runs at inputs $\{u^n\}_{n=1}^N$ to obtain outputs $\{f(u^n)\}_{n=1}^N$, and then follow the Bayesian procedure outlined in the previous section to obtain the predictive process $f_N \sim \text{GP}(m^f_N(u), k_N(u,u'))$. As such, it is similar to many other surrogate models based on model runs or {\em snapshots}. 

The Gaussian process $f_N$ is a random surrogate model for $f$. The uncertainty in $f_N$, encoded in the predictive variance $k_N$, is a model for the remaining uncertainty about $f$ after observing the finite amount of information $\{u^n, f(u^n)\}_{n=1}^N$, or in other words a model for the error between $f$ and $m^f_N$, based on the assumption that $f$ is a sample of the Gaussian process prior \eqref{eq:gp}. At the training points $D_N$, where we have observed $f$, there is no uncertainty about what value $f$ takes and there is no error in $m_N^f$ (i.e. $k_N(u^n, u^n) = 0$ and $m_N^f(u^n) = f(u^n)$, cf Figure \ref{fig:samples_posterior}). Away from the training points, where we do not know the value of $f$, there is still some uncertainty/error (i.e. $k_N(u, u) > 0$ and $m_N^f(u) \neq f(u)$ in general).

It remains to see that $f_N$ is indeed a surrogate model in the sense that it is much cheaper to evaluate than the original model $f$. To this end, note that in the approximate posteriors in section \ref{ssec:inv_surrogate}, we typically need to evaluate the predictive mean $m_N^f$ and the predictive variance $k_N$. As can be seen from \eqref{eq:pred_eq_mean}, the predictive mean $m_N^f$ is a linear combination of kernel evaluations:
\begin{equation*}
m_N^f(u) = \sum_{n=1}^N \alpha_n k(u,u^n), \qquad \alpha = K(D_N,D_N)^{-1} f(D_N) \in \R^N.
\end{equation*}
To evaluate $m_N^f(u)$ at an unobserved input $u \in U \setminus D_N$, we hence only need to evaluate the sum on the RHS of the expression above. Firstly, we note that the number of summands $N$ is typically small. This corresponds to the number of times we need to run our original model $f$, and in many practical applications, this will be limited by computational resources. Secondly, we note that the computation of the coefficients $\alpha$ can, {in general, be somewhat costly, since it involves finding the Cholesky factorization (or similar) of $K(D_N,D_N)$ which incurs computational cost $\mathcal O(N^3)$; however, as noted previously, $N$ is typically small in the applications of surrogate modeling that we consider in this paper.} Furthermore, the coefficients $\alpha$ only need to be computed {\em once}, rather than for every $u$. Thirdly, we note that the evaluation of the kernel $k(u,u^n)$ is cheap, since this typically involves the evaluation of standard functions such as polynomials and exponentials. Similar arguments apply to evaluations of the predictive variance $k_N(u,u)$.


\subsection{Examples of Gaussian Process Regression as Surrogate Model}
\label{ssec:exsm}

{We return to the two examples introduced in subsection \ref{ssec:exbip}
in order to illustrate the potential speedups afforded by use of
Gaussian process surrogate modeling.}

\begin{example} {Computational Timings in Subsurface Flow Example} Consider the subsurface flow example from subsection \ref{ssec:exbip}. The table below gives representative computational timings comparing cost of the evaluation of the
solution of the PDE, $f(u)$, with the two primary costs incurred in the Gaussian
process surrogate modeling, namely the determination of the coefficients of the kernel representation of the mean, $\alpha$, and evaluation of the mean $m_N^f$. Further details, and more extensive numerical experiments, may be found in \cite{btz23}. The numerical solution of the PDE for a given parameter $u$, representing the true model $f$, is implemented in Firedrake \cite{firedrake2016}.

\begin{table}[h!]
\centering
\begin{tabular}{|c | c c c|} 
 \hline
 Set-up & Computing $f(u)$ & Computing $m_N^f(u)$ & Computing $\alpha$ \\ [0.5ex] 
 \hline\hline
 $d_u = 2, d_y=10, D = (0,1), N=10$ & $2.6 \times 10^{-1}$s & $ 3.6 \times 10^{-5}$s & $3.2 \times 10^{-3}$s \\ 
 $d_u = 2, d_y=10, D = (0,1), N=20$ & $2.6 \times 10^{-1}$s & $ 4.0 \times 10^{-5}$s & $1.2 \times 10^{-2}$s \\
  $d_u = 10, d_y=10, D = (0,1), N=10$ & $2.6 \times 10^{-1}$s & $ 5.6 \times 10^{-5}$s & $3.2 \times 10^{-3}$s \\ 
[1ex] 
 \hline
\end{tabular}
\label{table:1}
\end{table}

\end{example}

\begin{example} {Computational Motivation for GCM Example}
In the paper \cite{dunbar2021calibration} Gaussian process regression is
used to accelerate Bayesian inference for parameters appearing in an idealized general
circulation model (GCM) used in climate modeling and described in subsection
\ref{ssec:exbip}. Evaluation of the 
parameter-to-data map is computationally
expensive and use of Gaussian process surrogates, as
overviewed in this paper, can be used to accelerate the computations. In
particular the number of unknown parameters in the example considered in 
\cite{dunbar2021calibration} is $2$, the number of observations is on the order of
${\mathcal O}(10^2)$, whilst the number of variables in the computational model
is on the order of ${\mathcal O}(10^5)$; by using Gaussian process surrogates, which
sidestep the need to work in space of dimension ${\mathcal O}(10^5)$, computational
costs can be significantly reduced. The natural measure
of computational cost is the number of evaluations of the GCM required to perform MCMC to obtain
solution of the Bayesian inverse problem. Without emulation this is $\mathcal{O}(10^5)$; with
Gaussian process emulation it is possible to achieve the same accuracy in solution of the
Bayesian inverse problem with $\mathcal{O}(10^2)$ evaluations of the forward model \cite{dunbar2021calibration}. These evaluations arise from providing training data for the Gaussian process and in particular from the use of the ensemble Kalman based calibrate-emulate-sample approach to the design of this training data \cite{cleary2021calibrate}. The wish to put the
calibrate-emulate-sample methodology on firm theoretical foundations motivates our work
in this paper on the optimal choice of design points for Gaussian process surrogate modeling
in the context of Bayesian inverse problems.
\end{example}

\subsection{Error in GP-accelerated Posterior Distribution}
\label{ssec:error0}

Suppose now that we use Gaussian process regression as our surrogate model in the approximate posterior distributions \eqref{eq:rad_nik_mean} and \eqref{eq:rad_nik_marginal}. An application of Theorems \ref{thm:mean_posterior} and \ref{thm:marginal_posterior} gives the following error bounds. We denote by $H_k(U)$ the reproducing kernel Hilbert space (RKHS) (see e.g. \cite{wendland}) of the prior covariance kernel $k$ in \eqref{eq:gp}.

\begin{corollary} \label{cor:post_conv_gp_phi} Suppose $U \subseteq \R^{d_u}$, $\Phi \in H_k(U)$, and the random surrogate model is constructed by applying Gaussian process regression to $\Phi$, resulting in $\Phi_N \sim \text{GP}(m^\Phi_N(u), k_N(u,u'))$. Then there exist constants $C_{\mathrm{Cor} \ref{cor:post_conv_gp_phi}} , C_{\mathrm{Cor} \ref{cor:post_conv_gp_phi}} ' > 0$, independent of $N$, such that \vspace*{-1ex}
\begin{align*}
\dhh(\mu^y, \mu^{y,N}_\mathrm{mean}) &\leq C_{\mathrm{Cor} \ref{cor:post_conv_gp_phi}}  \left\|\Phi - m^\Phi_N \right\|_{L^{2}_{\mu^y}(U)}, \end{align*}
and, under the additional assumptions that $U$ is bounded with Lipschitz boundary and $\sup_{u \in U} k_N(u,u) \rightarrow 0$ as $N \rightarrow \infty$, \vspace*{-1ex}
\begin{align*}
\dhh(\mu^y, \mu^{y,N}_\mathrm{marginal}) &\leq C_{\mathrm{Cor} \ref{cor:post_conv_gp_phi}} ' \left( \left\|\Phi - m^\Phi_N\right\|_{L^{2}_{\mu^y}(U)} + \left\| k_N^{1/2}(\cdot, \cdot) \right\|_{L^{2}_{\mu^y}(U)} \right).
\end{align*}
\end{corollary}

\begin{corollary} \label{cor:post_conv_gp_G} Suppose $U \subseteq \R^{d_u}$, $\mathcal G^j \in H_k(U)$ for $j=1,\dots,d_y$, and the random surrogate model is constructed by applying Gaussian process regression component-wise to $\mathcal G$, resulting in $\mathcal G_N^j \sim \text{GP}(m^{\mathcal G^j}_N(u), k_N(u,u'))$ and $\Phi_N(u) =  \frac{1}{2} \left\| y -  \mathcal G_N(u) \right\|_{\Gamma}^2$. Then there exist constants $C_{\mathrm{Cor}\ref{cor:post_conv_gp_G}} , C_{\mathrm{Cor}\ref{cor:post_conv_gp_G}}' > 0$, independent of $N$, such that \vspace*{-1ex}
\begin{align*}
\dhh(\mu^y, \mu^{y,N}_\mathrm{mean}) &\leq C_{\mathrm{Cor}\ref{cor:post_conv_gp_G}} \sum_{j=1}^{d_y} \left\|\mathcal G^j - m^{\mathcal G^j}_N \right\|_{L^{2}_{\mu^y}(U)},
\end{align*}
and, under the additional assumptions that $U$ is bounded with Lipschitz boundary and $\sup_{u \in U} k_N(u,u) \rightarrow 0$ as $N \vspace{-0.5ex}\rightarrow \infty$, \vspace*{-1ex}
\begin{align*}
\dhh(\mu^y, \mu^{y,N}_\mathrm{marginal}) &\leq C_{\mathrm{Cor}\ref{cor:post_conv_gp_G}}' \left( \sum_{j=1}^{d_y}  \left\|\mathcal G^j - m^{\mathcal G^j}_N \right\|_{L^{2}_{\mu^y}(U)} + \left\| k_N^{1/2}(\cdot, \cdot) \right\|_{L^{2}_{\mu^y}(U)} \right).
\end{align*}
\end{corollary}

Proofs of the above results are similar to those in \cite{st18} developed 
using the $L^2$-norm weighted by the prior measure $\mu_0$, $L^{2}_{\mu_0}(U).$ 
The proofs are outlined in the appendix.
However, in contrast to \cite{st18}, for the new results derived here, all error measures are 
computed in the $L^2$-norm weighted by the true posterior measure $\mu^y$,
$L^{2}_{\mu^y}(U).$ This enables us to deduce that the behaviour of the random surrogate model $\Phi_N$ in parts of the parameter space to which $\mu^y$ does not attach significant mass does not have a big influence on the accuracy of the approximate posterior distributions.

For the mean-based approximation \eqref{eq:rad_nik_mean} the results
show that we only need to control the error in the predictive means $m^\Phi_N$ and $m^{{\mathcal G}^j}_N$, respectively. For the marginal approximations \eqref{eq:rad_nik_marginal}, we also need to control the predictive variance $k_N$.

\subsection{Error Bounds for GP Regression in Weighted Spaces}
\label{ssec:gp_sk}
We now study the behaviour of the predictive mean $m_N^f$ and predictive variance $k_N$ as $N \rightarrow \infty$, for a general function $f : U \rightarrow \R$ that we wish to emulate. Using a representer theorem (\cite{rasmussen_williams,schoelkopf2002learning}), 
{the predictive mean $m_N^f$ defined in \eqref{eq:pred_eq_mean}} can be shown to be the minimum norm interpolant in the RKHS $H_k(U)$ of the prior covariance kernel $k$:
\begin{equation}\label{eq:mean_min} 
m_N^f = \argmin_{\substack{{g \in H_k(U) \; \text{s.t. }} \\ {g(u^{(n)}) = f(u^{(n)}), \, 1 \leq n \leq N}}} \|g\|_{H_k(U)}.
\end{equation}
For the Mat\`ern kernels defined in \eqref{eq:mat_cov}, it is known that the RKHS is equal to the Sobolev space $H^{\nu + d_u/2}(U)$ as a vector space, with equivalent norms \cite{wendland}. So there exist constants $C_\mathrm{low}(U), C_\mathrm{up}(U) > 0$ such that for all $g \in H_k(U)$, we have
\begin{equation}\label{eq:norm_equi}
C_\mathrm{low}(U) \|g\|_{H^{\nu + d_u/2}(U)} \leq \|g\|_{H_{k}(U)}  \leq C_\mathrm{up}(U) \|g\|_{H^{\nu + d_u/2}(U)}. 
\end{equation}
{Given the training points $D_N$ we introduce the mapping $m_N^{(\cdot)} : H_k(U) \rightarrow H_k(U)$, which is built on the definition of the predictive mean $m_N^f$ given in \eqref{eq:pred_eq_mean}. With $g(D_N) := [g(u^1); \dots; g(u^N)] \in \R^N$ as in \eqref{eq:data_exact}, we let
\begin{align} \label{eq:pred_mean_g}
g(u) \mapsto m_N^g(u) = k(u,D_N)^T K(D_N,D_N)^{-1} g(D_N).
\end{align}
The predictive variance $k_N$ satisfies the equality in Proposition \ref{prop:predvar_sup}, which follows, allowing us to transfer convergence results on $m_N^f$ to convergence results on $k_N$ \cite{sss13,st18}.

\begin{proposition}\label{prop:predvar_sup} Suppose $k_N$ and $m_N^{(\cdot)}$ are given by \eqref{eq:pred_eq_var} and \eqref{eq:pred_mean_g}, respectively. Then for any $u \in U$ we have
\[
k_N(u,u)^{\frac{1}{2}} = \sup_{\substack{{g \in H_k(U) \; \text{s.t. }} \\ \|g\|_{H_k(U)}=1}} | g(u) - m^g_N(u)|.
\]
\end{proposition}}
For any bounded set $\tilde U \subseteq U$, we define the {fill distance} $h_{D_N, \tilde U}$ as
\begin{equation*}
h_{D_N, \tilde U} := \sup_{u \in \tilde U} \inf_{u^n \in D_N \cap \tilde U} \|u- u^n\|_2.
\end{equation*}
The fill distance is the maximum distance any point in $\tilde U$ can be from a training point $u^n \in D_N \cap \tilde U$, and we require a space-filling set of training points $D_N$ (in $\tilde U$) to ensure that the fill distance goes to zero as $N \rightarrow \infty$. The following result follows from standard results in scattered data approximation \cite{wendland,nww05,alt12}.

\begin{proposition}\label{prop:scat_zeros} Suppose $\tilde U \subseteq U$ is a Lipschitz domain that satisfies an interior cone condition with angle $\theta$, and $\tilde U$ is contained in the cube $B(u_c, R_c) = \{ u \in \R^{d_u} : \|u - u_c\|_\infty \leq R_c \}$, for some $u_c \in \R^{d_u}$ and $0 < R_c < \infty$. Suppose further that the RKHS $H_k(\tilde U)$ is isomorphic to the Sobolev space $H^\tau(\tilde U)$ and $f \in H^\tau(\tilde U)$. Then there exist constants $C_{\mathrm{Prop} \ref{prop:scat_zeros}}, C_{\mathrm{Prop} \ref{prop:scat_zeros}}' > 0$, independent of $f$, $D_N$ and $N$, such that
\[
\| f - m_N^f\|_{H^\beta(\tilde U)} \leq C_{\mathrm{Prop  }\ref{prop:scat_zeros}}(\theta) h_{D_N, \tilde U} ^{\tau - \beta} \|f\|_{H^\tau(U)}, \qquad \text{for any } \beta \leq \tau,
\]
for all sets $D_N$ with $h_{D_N, \tilde U}  \leq h_0(\tilde U) = C_{\mathrm{Prop} \ref{prop:scat_zeros}}' R_c$. 
\end{proposition}


{Proposition \ref{prop:scat_zeros} allows us to bound the error $f - m_N^f$ locally in a subdomain $\tilde U$, in the sense that $\| f - m_N^f\|_{H^\beta(\tilde U)}$ can be bounded in terms of the local fill distance $h_{D_N, \tilde U}$. The proof is given in the appendix. }

The behaviour of the fill distance $h_{D_N, \tilde U}$ in terms of $N$ can be characterized explicitly for many point sets $D_N$, see e.g.  \cite{t20,wbg21} and the references therein. The fastest possible decay as $N \rightarrow \infty$ is $h_{D_N, \tilde U} = \mathcal O(N^{-\frac{1}{d_u}})$, which is obtained for example by uniform tensor grids. As seen in Proposition \ref{prop:rand_fill} below, the fill distance of randomly sampled point sets also decays at (almost) the optimal rate.

\begin{proposition} \label{prop:rand_fill} Suppose 
\begin{enumerate}
\item[(i)] $\tilde U \subseteq \mathbb R^{d_u}$ is a bounded Lipschitz domain that satisfies an interior cone condition, and $\tilde U$ is contained in the cube $B(u_c, R_c) = \{ u \in \R^{d_u} : \|u - u_c\|_\infty \leq R_c \}$, for some $u_c \in \R^{d_u}$ and $0 < R_c < \infty$,
\item[(ii)] $g : [0, \infty) \rightarrow [0, \infty)$ is continuous, monotonically increasing, and satisfies $g(0) = 0$ and $\lim_{x \downarrow 0} g(x) \exp(x^{-3d_u})= \infty$,
\item[(iii)] training points $\tilde D_N \subseteq \tilde U$ are sampled i.i.d. from a measure $\nu$ with density $\rho$ satisfying $\rho(u) \geq \rho_\mathrm{min} > 0$ for all $u \in \overline{\tilde U}$. 
\end{enumerate}
Then there exists constants $C_{\mathrm{Prop } \ref{prop:rand_fill} }, C_{\mathrm{Prop } \ref{prop:rand_fill} }' > 0$ and  $0 < C_{\mathrm{Prop } \ref{prop:rand_fill} }'' \leq 1$, independent of $N$, $u_c$ and $R_c$, such that for any $\varepsilon > 0$, we have
\begin{align*}
\EE_\nu[g( h_{\tilde D_N, \tilde U})] &\leq C_{\mathrm{Prop } \ref{prop:rand_fill}}\, R_c \, g (N^{-\frac{1}{d_u} + \varepsilon}), \\
\PP_\nu[h_{\tilde D_N, \tilde U} > h] &\leq C_{\mathrm{Prop } \ref{prop:rand_fill}}'  \, \left( \frac{h}{R_c}\right)^{-d_u} \left( 1 - C_{\mathrm{Prop } \ref{prop:rand_fill}}''  \left( \frac{h}{R_c}\right)^{d_u} \right)^{N}, \quad \textrm{for any } 0 \leq h \leq R_c.
\end{align*}
\end{proposition}

The proof of Proposition \ref{prop:rand_fill} in the special case $\tilde U \subseteq [0,1]^{d_u}$ can be found in \cite{ocbg19}. The general case follows from a simple transformation $u \mapsto R_c u + u_c$. The constants $C_{\mathrm{Prop } \ref{prop:rand_fill}}$ and $C_{\mathrm{Prop } \ref{prop:rand_fill}}''$ deteriorate, to $\infty$ and $0$ respectively, as $\rho_\mathrm{min} \rightarrow 0$.
The results of Proposition \ref{prop:rand_fill} further extend to the setting where the training points are not sampled i.i.d. from $\nu$, but are instead generated from a uniformly ergodic Markov chain with $\nu$ as its stationary distribution \cite{ocbg19}. 

Motivated by the results in Corollaries \ref{cor:post_conv_gp_phi} and \ref{cor:post_conv_gp_G}, we now study the quantitites $\EE_\nu \left[  \|f - m_N^f\|_{L^2_{\mu^y}(U)} \right]$ and $\EE_\nu \left[  \|k_N^{1/2}(\cdot, \cdot)\|_{L^2_{\mu^y}(U)}\right]$, where the design points $D_N$ are assumed to be sampled from $\nu$. We are particularly interested in the interplay between the sampling measure $\nu$ (with density $\rho$) and the posterior $\mu^y$ (with density $\pi^y$). Intuitively, we would expect $\nu \approx \mu^y$, since Gaussian process regression will be more accurate in regions with a higher density of training points.

\begin{theorem} \label{thm:gp_conv_weighted} Suppose $U$ is a bounded Lipschitz domain, $H_k(U)$ is isomorphic to the Sobolev space $H^\tau(U)$, and $ f \in H^\tau(U)$. Further suppose that for all $N \in \N$,
\begin{enumerate}
\item[(i)] $U_N \subseteq \R^{d_u}$ is compact and $U_N \subseteq \left\{u \in \R^{d_u} : \pi^y(u) \leq C_1^2 N^{-\frac{2 \tau}{d_u}} \right\}$,
\item[(ii)] the training points $D_N$ are sampled i.i.d. from a measure $\nu_N$ with density $\rho_N$ satisfying $\rho_N(u) \geq \rho_{\mathrm{min}} > 0$ for all $u \in \overline{U \setminus U_N}$, and $\rho_N(u) = 0$ otherwise,
\item[(iii)] $U \setminus U_N $ is a Lipschitz domain that satisfies an interior cone condition with angle $\theta$, and $U \setminus U_N $ is contained in the cube $B(u_c, R_c^N) = \{ u \in \R^{d_u} : \|u - u_c\|_\infty \leq R_c^N \}$, for some $u_c \in \R^{d_u}$ and $0 < R_c^N < C_2 \log N$.
\end{enumerate} 
Then there exists a constant $C_{\mathrm{Thm } \ref{thm:gp_conv_weighted}} >0$, independent of $f$ and $N$, such that for all $0 \leq \beta \leq \tau$ and $\varepsilon > 0$ we have
\[
\EE_{\nu_N} \left[  \|f - m_N^f\|_{H^\beta_{\mu^y}(U)} \right] \leq C_{\mathrm{Thm } \ref{thm:gp_conv_weighted}} \; N^{-\frac{\tau-\beta}{d_u} + \varepsilon} \, \|f\|_{H^\tau(U)}.
\]
Furthermore, for any partitioning $U \setminus U_N \subseteq \cup_{i=1}^r B_i$, where each $B_i$ is a bounded Lipschitz domain that satisfies an interior cone condition with angle $\theta'$, there exists a constant $C_{\mathrm{Thm } \ref{thm:gp_conv_weighted}}' $ such that for all $0 \leq \beta \leq \tau$ we have
\begin{align*}
&\EE_{\nu_N} \left[  \|f - m_N^f\|_{H^\beta_{\mu^y}(U)} \mathrm{I}_{\{ h_{D_N, B_i}  \, \leq \, h_0(B_i), 1 \leq i \leq n \}} \right] \\
&\leq C_{\mathrm{Thm } \ref{thm:gp_conv_weighted}}'  \left( \left(\sup_{u \in U_N} \pi^y(u)\right)^{\frac{1}{2}} + \sum_{i=1}^{r} \left( \sup_{u \in B_i} \pi^y(u) \right)^{\frac{1}{2}}  \, \EE_{\nu_N} \left[ h_{D_N, B_i}^{\tau - \beta} \right] \right).
\end{align*}

\end{theorem}

The choice $\beta = 0$ in Theorem \ref{thm:gp_conv_weighted} gives a bound on $\|f - m_N^f\|_{L^2_{\mu^y}(U)}$, whereas $\beta = d_u/2 + \delta$, for any $\delta > 0$, gives a bound on $\|k_N^{1/2}(\cdot, \cdot)\|_{L^2_{\mu^y}(U)}$ via Proposition \ref{prop:predvar_sup} and the Sobolev embedding theorem as in \cite{st18}. The assumption that $U$ is bounded is required only since we need $\rho_\mathrm{min} > 0$ in assumption $(ii)$. The assumption that $R_c^N \leq C_2 \log N$ in $(iii)$ can be dropped since $U$ is bounded and hence $R_c^N$ is uniformly bounded in $N$. However, $R_c^N$ does appear as a factor in $C_{\mathrm{Thm } \ref{thm:gp_conv_weighted}}$ and so different choices of $R_c^N$ could lead to pre-asymptotic effects slowing down convergence. In assumption $(iii)$ we assume that the angle $\theta$ is independent of $N$, which ensures that the constant $C_{\mathrm{Prop  }\ref{prop:scat_zeros}}(\theta)$ in Proposition \ref{prop:scat_zeros} is also independent of $N$.

The results in Theorem \ref{thm:gp_conv_weighted} give us insight into the interplay between the choice of training points $D_N$ and the posterior $\mu^y$. We introduce the set $U_N$ as the part of the parameter space $U$ which does not carry significant posterior mass. We then place training points only in $U \setminus U_N$, according to a sampling measure $\nu_N$. The first claim shows that even by ignoring the set $U_N$, and placing training points only where the posterior density is sufficiently large, we still obtain (almost) the optimal rate of convergence in $N$. Optimality here refers to the fastest obtainable rate $N^{-1/d_u}$ of the fill distance in $d_u$ dimensions, as well as the fastest obtainable rate $N^{-\tau/d_u}$ for the approximation of $f \in H^\tau(U)$ in $L^2(U)$ by $N$ function values (see e.g. \cite{t20} and the references therein). The first claim holds for any choice of sampling measure $\nu_N$ on $U \setminus U_N$, but in general we expect $C_{\mathrm{Thm } \ref{thm:gp_conv_weighted}}$ to deteriorate as $\rho_{ \mathrm{min}}$ gets close to zero.

The second claim shows that the sampling measure $\nu_N$ should be chosen such that the local fill distance $h_{D_N, B_i}$ is small where the posterior mass $\sup_{u \in B_i}\pi^y(u)$ is large. In other words, we want to put training points in the regions with highest posterior density.
We also note that the bound in the second claim suggests oversampling in the tails of the posterior density $\pi^y$, since the fill distance should be balanced with the {\em square-root of $\pi^y$}. This might be related to the fact that extrapolation is generally much harder than interpolation.

Finally, we remark that the above observations agree with previous studies on the subject. In \cite{sn17}, the authors develop a sequential strategy for optimally choosing the training points $D_N$, in the context of the marginal approximation \eqref{eq:rad_nik_marginal}. In practice this results in the training points being placed in regions of high posterior density. Similarly, the choice of design points attained by the calibrate-emulate-sample approach in \cite{cleary2021calibrate,dunbar2021calibration}, gives training points in regions of high posterior density for the solution of the Bayesian inverse problem in the GCM example from subsection \ref{ssec:exbip}. The results in this paper are the first to provide a theoretical justification for this choice of training points in terms of accuracy of the approximate posterior distributions.

The related problem of optimal choice of training points $D_N$ in Bayesian quadrature, i.e. when we are interested in bounding the error 
\[
\left| \int_U f(u) \pi(u) \mathrm{d} u - \int_U m_N^f(u) \pi(u) \mathrm{d} u \right|,
\]
for some target measure $\pi$, is studied numerically in \cite{boccg17}. They observe that choosing the design points $D_N$ according to a slightly inflated version of $\pi$, i.e. oversampling in the tails of $\pi$, seems to give the smallest error.


\section{Numerical Examples}\label{sec:num}
We finish this work with two simple illustrative examples. In the first example,  our function is $f(u)=u$, and our posterior measure is $\mu \sim \mathcal{N}(1,1)$. We then take our design measure to be $\nu \sim \mathcal{N}(1,\sigma^{2})$. We are interested in analyzing  the following error quantity
\begin{equation} \label{eq:error}
e(N,\nu)=\int_{-\infty}^{\infty} \mathbb{E}_{\nu}| m^f_{N}(u)-f(u)|^{2}\mu(du)
\end{equation}
both as a function of the value of the variance $\sigma^{2}$ of the design measure $\nu$ as well as the number of points $N$ used in the Gaussian process regression. In all our calculations we have used $10^{3}$ realizations for the design points in order to approximate the expectation in \eqref{eq:error}. Note that this example does not fulfill all assumptions of Theorem \ref{thm:gp_conv_weighted}, since $U$ is unbounded and there is no a-priori truncation of the parameter space by discarding $U_N$, but the behaviour is still as {predicted by that theorem}.

\begin{figure}
\begin{center}
\subfloat{\includegraphics[width=.45\textwidth]{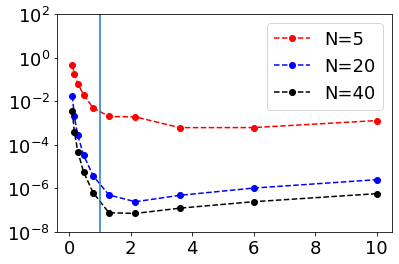}}\quad
  \subfloat{\includegraphics[width=.45\textwidth]{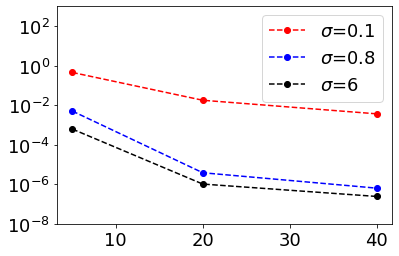}}\\
\end{center}
\caption{$e(N,\nu)$ as a function of $\sigma$ and $N$ for the case of Gaussian posterior $\mu$ and design measure $\nu$, with Gaussian covariance kernel $k$.}
\label{fig:comparison_gaussian}
\end{figure}

As we can in Figure \ref{fig:comparison_gaussian}, the results agree with the Theorem \ref{thm:gp_conv_weighted}. In particular, as expected we observe that as the number of design points $N$ increases $e(N,\nu)$ decreases, and at the same rate for different choices of $\sigma$. Furthermore, perhaps more interestingly we observe that as we increase the variance of the design measure the error overall decreases until it reaches a minimum value nearby the value of the variance of the true posterior and then slowly increases.

\begin{figure}
\begin{center}
\subfloat{\includegraphics[width=.45\textwidth]{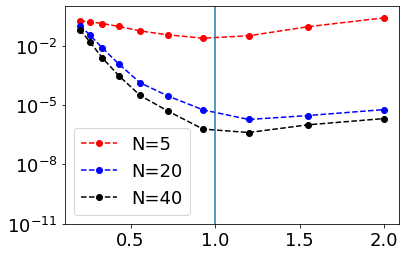}}\quad
  \subfloat{\includegraphics[width=.45\textwidth]{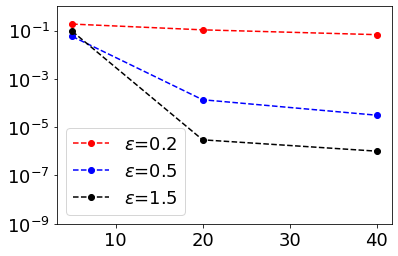}}\\
\end{center}
\caption{$e(N,\nu)$ as a function of $\epsilon$ and $N$ for the case of uniform posterior $\mu$ and design measure $\nu$, with Gaussian covariance kernel $k$.}
\label{fig:comparison_uniform}
\end{figure}

{In the second numerical experiment} we now repeat the first experiment, again with $f(u)=u$, but now with the true posterior measure $\mu \sim U[-1,1]$ and the design measure $\nu \sim U[-\epsilon,\epsilon]$. As we can see in  Figure \ref{fig:comparison_uniform} the results are similar to the Gaussian case studied before with the error reducing as a function of $N$ for a fixed value of $\epsilon$.

\section*{Acknowledgements}
ALT would like to thank the Isaac Newton Institute for Mathematical Sciences, Cambridge, for support and hospitality during the programme {\em Mathematical and statistical foundations of future data-driven engineering} where work on this paper was undertaken. This work was supported by EPSRC grant no EP/R014604/1.
AMS is also grateful to the National Science Foundation (grant
AGS-1835860).

\bibliographystyle{siam}
\bibliography{bibgp}

\section*{Appendix: Proofs of Results}
\addcontentsline{toc}{section}{Appendix: Proofs of Results}
\setcounter{equation}{0}
\renewcommand{\theequation}{A.\arabic{equation}}

\begin{proof}[of Theorem \ref{thm:mean_posterior}]
First, note that the mean-based posterior approximation $\mu^{y,N}_\mathrm{mean}$ is absolutely continuous with respect to the true posterior $\mu^y$, since $\exp\big(-\EE [\Phi_N(u)]\big)$ and $\exp\big(- \Phi(u) \big)$ are both positive and $\mu_0$ is a probability measure:
\[
\mu^y(A) = \int_A \exp\big(- \Phi(u) \big) \mu_0(\mathrm{d}u) = 0 \Rightarrow \mu^{y,N}_\mathrm{mean}(A) = \int_A \exp\big(-\EE [\Phi_N(u)]\big) \mu_0(\mathrm{d}u) = 0.
\]
For the same reasons, the prior $\mu_0$ is absolutely continuous with respect to the true posterior $\mu^y$. We then have the Radon-Nikodym derivative
\[
\frac{d\mu^{y,N}_\mathrm{mean}}{d\mu^y}(u) = \frac{d\mu^{y,N}_\mathrm{mean}}{d\mu_0}(u) \frac{d\mu_0}{d\mu^y}(u) = \frac{d\mu^{y,N}_\mathrm{mean}}{d\mu_0}(u) \left( \frac{d\mu^y}{d\mu_0}(u)\right)^{-1} = \frac{ \frac{1}{Z_{N}^\mathrm{mean}}  \exp\big(-\EE [\Phi_N(u)]\big)}{\frac{1}{Z}  \exp\big(- \Phi(u) \big)} .
\]
 
Using the definition of the Hellinger distance with $\nu = \mu^y$ and the inequality
\[
\left(1-\frac{a b}{c d } \right)^2= \frac{1}{c^2 d^2} (cd - ab)^2 = \frac{1}{c^2 d^2} (cd - cb + cb- ab)^2 \leq \frac{2}{c^2 d^2} \left( (cd- cb)^2 + (cb- ab)^2 \right),
\]
for real numbers $a,b,c,d \in \R$, we then have
\begin{align*}
2  \; \dhh^2(\mu^y, \mu^{y,N}_\mathrm{mean}) &=\int_U \left( 1 - \sqrt{\frac{ \frac{1}{Z_{N}^\mathrm{mean}}  \exp\big(-\EE [\Phi_N(u)]\big)}{\frac{1}{Z}  \exp\big(- \Phi(u) \big)}} \right)^2 \mu^y(\mathrm{d}u) \\
&\leq 2 \int_U \left(1 - \exp\left(\frac{1}{2} (\Phi(u) - \EE [\Phi_N(u)])\right)\right)^2 \mu^y(\mathrm{d}u) \; + \\
&\qquad \qquad 2 \, Z \left(Z^{-1/2} - (Z_{N}^\mathrm{mean})^{-1/2} \right)^2 \int_U \exp\big((\Phi(u) - \EE [\Phi_N(u)])\big) \mu^y(\mathrm{d}u)\\
&=: I + II.
\end{align*}

Using the local Lipschitz continuity of the exponential function, H\"older's inequality with conjugate exponents $(p_1,p_1')$, the inequality $(a+b)^2 \leq 2(a^2 + b^2)$ for $a,b \in \R$, the triangle equality in $L^{p_1}_{\mu^y} $, and Jensen's inequality for $x \rightarrow x^2$, we have
\begin{align*}
I &\leq \frac{1}{2} \int_U \left(1 + \exp\left(\frac{1}{2} (\Phi(u) - \EE [\Phi_N(u)])\right)\right)^2 \left(\Phi(u) - \EE [\Phi_N(u)] \right)^2 \mu^y(\mathrm{d}u) \\
&\leq \frac{1}{2} \left\| \left(1 + \exp\left(\frac{1}{2} (\Phi - \EE [\Phi_N])\right)\right)^2 \right\|_{L^{p_1}_{\mu^y}} \left\| \left(\Phi - \EE [\Phi_N] \right)^2 \right\|_{L^{p_1'}_{\mu^y}}  \\
&\leq ( 1+ C_1(p_1)) \left\| \Phi - \EE [\Phi_N] \right\|^2_{L^{2p_1'}_{\mu^y}}  
\end{align*}
For $II$, using the inequality $(a^{-1/2} - b^{-1/2})^2 \leq (a-b)^2 \max\{a^{-3}, b^{-3}\}$, for $a,b > 0$, we have
\begin{align*}
II &\leq 2 \, Z \max(Z^{-3},(Z_{N}^\mathrm{mean})^{-3}) \left(Z- (Z_{N}^\mathrm{mean})\right)^2  \int_U \exp\big((\Phi(u) - \EE [\Phi_N(u)])\big) \mu^y(\mathrm{d}u).
\end{align*}
Then, as in the bound for $I$ and using Jensen's inequality for $x \rightarrow x^2$, we have
\begin{align*}
\left(Z- (Z_{N}^\mathrm{mean})\right)^2 &= \left( \int_U \exp\big(- \Phi(u)\big) - \exp \big(- \EE [\Phi_N(u)])\big) \mu_0(\mathrm{d}u) \right)^2  \\
&= \left( \int_U \left(1 - \exp\left(\frac{1}{2} (\Phi(u) - \EE [\Phi_N(u)])\right)\right) \mu^y(\mathrm{d}u) \right)^2 \\
&\leq  ( 1+ C_1(p_1)) \left\| \Phi - \EE [\Phi_N] \right\|^2_{L^{2p_1'}_{\mu^y}}  
\end{align*}
and so
\begin{align*}
II &\leq 2 Z \, \max \left\{ Z^{-3}, C_2^3 \right\} (1+ C_1(p_1)) \, C_1(1) \, \left\| \Phi - \EE [\Phi_N] \right\|^2_{L^{2p_1'}_{\mu^y}}. 
\end{align*}
This completes the proof, with
\[
C_{\mathrm{Thm} \ref{thm:mean_posterior}} = \sqrt{ ( 1+ C_1(p_1)) \left(\frac{1}{2} +  Z \max \left\{ Z^{-3}, C_2^3 \right\} C_1(1) \right) }.
\]
\end{proof}

\begin{proof}[of Theorem \ref{thm:marginal_posterior}]
Similar to the proof of Theorem \ref{thm:mean_posterior}, we compute
\[
\frac{d\mu^{y,N}_\mathrm{marginal}}{d\mu^y}(u) = \frac{ \frac{1}{Z_{N}^\mathrm{marginal}} \EE \Big(\exp\big(-\Phi_N (u) \big)\Big)}{\frac{1}{Z}  \exp\big(- \Phi(u) \big)},
\]
and using the definition of the Hellinger distance with $\nu = \mu^y$, we have
\begin{align*}
2  \; \dhh^2(\mu^y, \mu^{y,N}_\mathrm{marginal}) =
&= \int_U \left( 1 - \sqrt{\frac{ \frac{1}{Z_{N}^\mathrm{marginal}} \EE \Big(\exp\big(-\Phi_N (u) \big)\Big)}{\frac{1}{Z}  \exp\big(- \Phi(u) \big)}} \right)^2 \mu^y(\mathrm{d}u) \\
&\leq 2 \int_U \left(1 - \sqrt{\EE \big[ \exp\big(\Phi(u) - \Phi_N(u)\big)\big]} \right)^2 \mu^y(\mathrm{d}u) \; + \\
&\qquad 2 \, Z \left(Z^{-1/2} - (Z_{N}^\mathrm{marginal})^{-1/2} \right)^2 \int_U \EE \big[ \exp\big(\Phi(u) - \Phi_N(u)\big)\big] \mu^y(\mathrm{d}u)\\
&=: I + II.
\end{align*}
Using the inequality $(a-b)^{2} = \left( \frac{a^{2}-b^{2}}{a+b} \right)^{2} \leq \frac{(a^{2}-b^{2})^{2}}{a^{2}+b^{2}} $, for $a,b \in \R$, together with H\"older's inequality with conjugate exponents $q_1$ and $q_1'$, we obtain
\begin{align*}
I &\leq 2 \left\| \left(1 - \EE \big[ \exp\big(\Phi - \Phi_N\big)\big] \right)^2 \right\|_{L^{q_1'}_{\mu^y}} \left\| \left(1 + \EE \big[ \exp\big(\Phi - \Phi_N\big)\big] \right)^{-1} \right\|_{L^{q_1}_{\mu^y}} .
\end{align*}
We estimate the second factor on the right-hand side above as in the proof of \cite[Theorem 3.1]{lst18}:
\[
\left\| \left(1 + \EE \big[ \exp\big(\Phi - \Phi_N\big)\big] \right)^{-1} \right\|_{L^{q_1}_{\mu^y}} \leq \min\left\{  \left\| 1\right\|_{L^{q_1}_{\mu^y}} ,  \left\| \EE \big[ \exp\big(\Phi - \Phi_N\big)\big]^{-1} \right\|_{L^{q_1}_{\mu^y}} \right\} = 1
\]
This holds for any $q_1 \in [1, \infty]$, and so we can choose $q_1 = \infty$ and $q_1'=1$. 

For the first factor, the linearity of expectation, local Lipschitz continuity of the exponential function, H\"older's inequality with conjugate exponents $p_2, p_2'$ with respect to $\EE$ and $p_1, p_1'$ with respect to $\mu^y$, the inequality $|a+b|^{p_2} \leq 2^{p_2-1}(|a|^{p_2} + |b|^{p_2})$ and the triangle inequality in $L^{2p_1}_{\mu^y}$ give \vspace*{-1ex}
\begin{align*}
I &\leq 2 \left\| \left(1 - \EE \big[ \exp\big(\Phi - \Phi_N\big)\big] \right)^2 \right\|_{L^{1}_{\mu^y}} \\
&\leq 2 \int_U \left( \EE \big[  \left(1 + \exp\left(\Phi(u) - \Phi_N(u)\right)\right) \left(\Phi(u) - \Phi_N(u) \right) \big] \right)^2 \mu^y(\mathrm{d}u) \\
&\leq 2 \left\| \EE \big[  \left(1 + \exp\left(\Phi - \Phi_N\right)\right)^{p_2} \big]^{1/{p_2}}  \right\|^2_{L^{2p_1}_{\mu^y}} \left\| \EE \big[ \left(\Phi - \Phi_N \right)^{p_2'} \big]^{1/p_2'} \right\|^2_{L^{2p_1'}_{\mu^y}}  \\
&\leq 2^{p_2} ( 1+ C_1(p_1,p_2)) \left\| \EE \big[ \left(\Phi - \Phi_N \right)^{p_2'} \big]^{1/p_2'} \right\|^2_{L^{2p_1'}_{\mu^y}} . 
\end{align*}
For $II$, using the inequality $(a^{-1/2} - b^{-1/2})^2 \leq (a-b)^2 \max\{a^{-3}, b^{-3}\}$, for $a,b > 0$, we have
\begin{align*}
II 
&\leq 2 \, Z \max(Z^{-3},(Z_{N}^\mathrm{marginal})^{-3}) \left(Z- (Z_{N}^\mathrm{marginal})\right)^2  \int_U \EE \big[ \exp\big(\Phi(u) - \Phi_N(u)\big)\big] \mu^y(\mathrm{d}u).
\end{align*}
Then, as in the bound for $I$ and using Jensen's inequality for $x \rightarrow x^2$, we have
\begin{align*}
\left(Z- (Z_{N}^\mathrm{marginal})\right)^2 &= \left( \int_U \exp\big(- \Phi(u)\big) - \EE \big[ \exp \big(- \Phi_N(u)\big) \big] \mu_0(\mathrm{d}u) \right)^2  \\
&= \left( \int_U \left(\EE \big[ 1 - \exp\big(\Phi(u) - \Phi_N(u)\big)\big] \right)^2 \mu^y(\mathrm{d}u) \right)^2 \\
&\leq  2^{p_2-1} ( 1+ C_1(p_1,p_2)) \left\| \EE \big[ \left(\Phi - \Phi_N \right)^{p_2'} \big]^{1/p_2'} \right\|^2_{L^{2p_1'}_{\mu^y}}  
\end{align*}
and so
\begin{align*}
II &\leq Z \, \max \left\{ Z^{-3}, C_2^3 \right\} \,2^{p_2} ( 1+ C_1(p_1,p_2)) C_1(1,1) \left\| \EE \big[ \left(\Phi - \Phi_N \right)^{p_2'} \big]^{1/p_2'} \right\|^2_{L^{2p_1'}_{\mu^y}} . 
\end{align*}
This completes the proof, with
\[
C_{\mathrm{Thm} \ref{thm:marginal_posterior}} = \sqrt{ 2^{p_2-1} ( 1+ C_1(p_1,p_2)) \left( 1+ Z \max \left\{ Z^{-3}, C_2^3 \right\} C_1(1,1) \right) }.
\]
\end{proof}

\begin{proof}[of Corollary \ref{cor:post_conv_gp_phi}] We check the assumptions in Theorems \ref{thm:mean_posterior} and \ref{thm:marginal_posterior}. For assumption $(ii)$, we can follow the proof of Lemmas 4.1 and 4.7 in \cite{st18}, respectively. Note that instead of assuming that $\sup_{u \in U} \left| \Phi(u) -  m^\Phi _N(u) \right|$ converges to 0 as $N$ tends to infinity and $\sup_{u \in U}\|\mathcal G(u)\| \leq C_\mathcal G < \infty$, we can directly bound $\sup_{u \in U} m_N^\Phi(u)$ independently of $N$ using the definition of the RKHS $H_k(U)$ of the kernel $k$, the Cauchy-Schwarz inequality and the minimum norm interpolant property \eqref{eq:mean_min}:
\begin{align*}
\sup_{u \in U}  \left| m_N^\Phi(u) \right| = \sup_{u \in U} \left| \langle m_N^\Phi(\cdot), k(\cdot, u) \rangle_{H_k(U)} \right| \leq \|\Phi\|_{H_k(U)} \sup_{u \in U} \sqrt{k(u, u)}.
\end{align*}
We can then bound $Z_{N}^\mathrm{mean}$ as in the proof of \cite[Lemma 4.1]{st18}, and $Z_{N}^\mathrm{marginal}$ as in the proof of \cite[Lemma 4.7]{st18}. For assumption (i) in Theorem \ref{thm:mean_posterior}, this then immediately follows with $p_1=\infty$, since as above
\begin{align*}
\left\| \exp\left((\Phi - \EE [\Phi_N])\right) \right\|_{L^{\infty}(U)} 
&\leq  \exp\left( \|\Phi\|_{H_k(U)} (1+\sup_{u \in U} \sqrt{k(u, u)}) \right).
\end{align*}
For assumption (i) in Theorem \ref{thm:marginal_posterior}, we similarly have that for $p_1=\infty$, the quantity
\[
\sup_{u \in U} \EE \big[  \left(\exp\left(\Phi(u) - \Phi_N(u) \right)\right)^{p_2} \big]^{1/{p_2}} \leq  \EE \big[ \exp\big(p_2 \sup_{u \in U} (\Phi(u) + \Phi_N(u) )\big) \big]^{1/{p_2}}
\]
can be bounded for any $p_2 < \infty$. The choice $p_2=2$ then gives the desired result.
\end{proof}

\begin{proof}[of Corollary \ref{cor:post_conv_gp_G}] This is very similar to the proof of Corollary \ref{cor:post_conv_gp_phi}, with the same modification to bound $\sup_{u \in U}  \big| m_N^{\mathcal G^j}(u) \big|$ independently of $N$.
\end{proof}

\begin{proof}[of Proposition \ref{prop:scat_zeros}] It follows from \cite[Theorem 3.2]{alt12} that 
\[
\| f - m_N^f\|_{H^\beta(\tilde U)} \leq C(\theta) h_{D_N, \tilde U} ^{\tau - \beta} \|f - m_N^f\|_{H^\tau(\tilde U)}, 
\]
for any $\beta \leq \tau$ and for all sets $D_N$ with $h_{D_N, \tilde U}  \leq h_0(\tilde U) = C_{\mathrm{Prop} \ref{prop:scat_zeros}}' R_c$. Using $\tilde U \subseteq U$, the triangle inequality in $H^\tau(U)$, the norm equivalence between $H_k(U)$ and $H^\tau(U)$ from \eqref{eq:norm_equi}, and $\|m_N^f\|_{H_k(U)} \leq \|f\|_{H_k(U)}$ from \eqref{eq:mean_min} then gives 
\begin{align*}
\|f - m_N^f\|_{H^\tau(\tilde U)} &\leq \|f \|_{H^\tau(U)} + \|m_N^f\|_{H^\tau(U)} \\
&\leq \left( 1 + C_\mathrm{low}(U)^{-1 } C_\mathrm{up}(U) \right) \|f\|_{H^\tau(U)}.
\end{align*}
This finishes the proof, with $C_{\mathrm{Prop  }\ref{prop:scat_zeros}}(\theta) = C(\theta) \left( 1 + C_\mathrm{low}(U)^{-1 } C_\mathrm{up}(U) \right)$.
\end{proof}

\begin{proof}[of Theorem \ref{thm:gp_conv_weighted}] \underline{Step 1:} We split the error as
\[
\EE_{\nu_N} \left[  \|f - m_N^f\|_{H^\beta_{\mu^y}(U)} \right] \leq \EE_{\nu_N} \left[  \|f - m_N^f\|_{H^\beta_{\mu^y}(U \setminus U_N)} \right] + \EE_{\nu_N} \left[  \|f - m_N^f\|_{H^\beta_{\mu^y}(U_N)} \right].
\]

\underline{Step 2:} For $\EE_{\nu_N} \left[  \|f - m_N^f\|_{H^\beta_{\mu^y}(U_N)} \right]$, we have by assumption $(i)$ and $\beta \leq \tau$ that
\begin{align*}
\EE_{\nu_N} \left[  \|f - m_N^f\|_{H^\beta_{\mu^y}(U_N)} \right] &\leq \left( \sup_{u \in U_N} \pi^y(u) \right)^{\frac{1}{2}}  \, \EE_{\nu_N} \left[ \|f - m_N^f\|_{H^\beta(U_N)}\right] \\
&\leq C_{1} N^{-\frac{\tau}{d_u}} \, \EE_{\nu_N} \left[  \|f - m_N^f\|_{H^\tau(U_N)} \right] \\
&\leq C_{1} \left( 1 + C_\mathrm{low}(U)^{-1 } C_\mathrm{up}(U) \right)  N^{-\frac{\tau}{d_u}} \, \|f\|_{H^\tau(U)},
\end{align*}
where we have bounded $\|f - m_N^f\|_{H^\tau(U_N)}$ as in the proof of Proposition \ref{prop:scat_zeros}.

\underline{Step 3:} For $\EE_{\nu_N} \left[  \|f - m_N^f\|_{H^\beta_{\mu^y}(U \setminus U_N)} \right]$, the linearity of expectation gives
\begin{align*}
\EE_{\nu_N} \left[ \| f - m_N^f\|_{H^\beta(U \setminus U_N)} \right] &= \EE_{\nu_N} \left[ \| f - m_N^f\|_{H^\beta(U \setminus U_N)} \mathrm{I}_{\{ h_{D_N, U \setminus U_N}  \leq h_0(U \setminus U_N) \}} \right] \\
&+ \EE_{\nu_N} \left[ \| f - m_N^f\|_{H^\beta(U \setminus U_N)} \mathrm{I}_{\{h_{D_N, U \setminus U_N}  > h_0(U \setminus U_N)} \} \right].
\end{align*}
For the first term, an application of Propositions \ref{prop:scat_zeros} and \ref{prop:rand_fill}, and H\"older's inequality with conjugate exponents $p=\infty$ and $q=1$, gives 
\begin{align*}
&\EE_{\nu_N} \left[ \| f - m_N^f\|_{H^\beta(U \setminus U_N)} \mathrm{I}_{\{ h_{D_N, U \setminus U_N}  \, \leq \, h_0(U \setminus U_N) \}} \right] \\
&\leq C_{\mathrm{Prop} \ref{prop:scat_zeros}}(\theta) \; \EE_{\nu_N} \left[ h_{D_N, U \setminus U_N}^{\tau - \beta} \right] \sup_{D_N} \|f - m_N^f\|_{H^\tau(U \setminus U_N)}  \\
&\leq C_{\mathrm{Prop} \ref{prop:scat_zeros}}(\theta) \; C_{\mathrm{Prop } \ref{prop:rand_fill}} \; C_2 \;\left( 1 + C_\mathrm{low}(U)^{-1 } C_\mathrm{up}(U) \right) \; N^{-\frac{\tau - \beta}{d_u} + \varepsilon} \|f\|_{H^\tau(U)},
\end{align*}
for any $\varepsilon > 0$, where we have bounded $\|f - m_N^f\|_{H^\tau(U \setminus U_N)}$ as in step 2. 

For the second term, we use H\"older's inequality with conjugate exponents $p=\infty$ and $q=1$ to obtain
\begin{align*}
&\EE_{\nu_N} \left[ \| f - m_N^f\|_{H^\beta(U \setminus U_N)} \mathrm{I}_{\{ h_{D_N, U \setminus U_N)}  \, > \,  h_0(U \setminus U_N)) \}} \right] \\
&\leq \sup_{D_N}  \| f - m_N^f\|_{H^\beta(U \setminus U_N)}  \PP_{\nu_N} \left[ h_{D_N, U \setminus U_N}  > h_0(U \setminus U_N) \right].
\end{align*}
The second factor can be bounded by Proposition \ref{prop:rand_fill}. For the first factor, we again use the bound on $\| f - m_N^f\|_{H^\beta(U \setminus U_N))}$ as in step 2.
Adding the two terms, we have 
\begin{align*}
&\EE_{\nu_N} \left[ \| f - m_N^f\|_{H^\beta(U \setminus U_N)} \right] \leq \left( 1 + C_\mathrm{low}(U)^{-1 } C_\mathrm{up}(U) \right) \|f\|_{H^\tau(U)} \\
& \left(   C_{\mathrm{Prop }\ref{prop:scat_zeros}}(\theta) \; C_{\mathrm{Prop } \ref{prop:rand_fill}} \, C_2 \, N^{-\frac{\tau - \beta}{d_u} + \varepsilon} +   C_{\mathrm{Prop } \ref{prop:rand_fill}}'  \; (C_{\mathrm{Prop }\ref{prop:scat_zeros}}')^{-d_u} \left( 1 - C_{\mathrm{Prop} \ref{prop:rand_fill}}''  (C_{\mathrm{Prop }\ref{prop:scat_zeros}}')^{d_u} \right)^{N} \right) .
\end{align*}
Since the geometrically decaying term is bounded by the algebraically converging term for $N$ sufficiently large, and both terms are monotonically decreasing in $N$, it follows that
\[
\EE_{\nu_N} \left[ \| f - m_N^f\|_{H^\beta(U \setminus U_N)} \right] \leq C N^{-\frac{\tau - \beta}{d_u} + \varepsilon} \|f\|_{H^\tau(U)}, 
\]
with $C = \left( 1 + C_\mathrm{low}(U)^{-1 } C_\mathrm{up}(U) \right) \max \{C_{\mathrm{Prop }\ref{prop:scat_zeros}}(\theta) C_{\mathrm{Prop } \ref{prop:rand_fill}} \, C_2,C_{\mathrm{Prop } \ref{prop:rand_fill}}'  \; (C_{\mathrm{Prop }\ref{prop:scat_zeros}}')^{-d_u}\}$. This proves the first claim, with $C_{\mathrm{Thm} \ref{thm:gp_conv_weighted} } = C + C_{1} \left( 1 + C_\mathrm{low}(U)^{-1 } C_\mathrm{up}(U) \right)$.

\underline{Step 5:} For the second claim, we split $\EE_{\nu_N} \left[  \|f - m_N^f\|_{H^\beta_{\mu}(U \setminus U_N)} \right]$ further over the subdomains $U \setminus U_N \subseteq \{B_i\}_{i=1}^{r} \subseteq U$ to obtain
\begin{align*}
&\EE_{\nu_N} \left[  \|f - m_N^f\|_{H^\beta_{\mu^y}(U \setminus U_N)} \mathrm{I}_{\{ h_{D_N, B_i}  \, \leq \, h_0(B_i), 1 \leq i \leq n \}} \right] \\
&\leq \sum_{i=1}^{r} \left( \sup_{u \in B_i} \pi^y(u) \right)^{\frac{1}{2}}  \, \EE_{\nu_N} \left[ \|f - m_N^f\|_{H^\beta(B_i)} \mathrm{I}_{\{ h_{D_N, B_i}  \, \leq \, h_0(B_i)} \right].
\end{align*}
Then, as in step 3, we have 
\begin{align*}
&\EE_{\nu_N} \left[ \| f - m_N^f\|_{H^\beta(B_i)} \mathrm{I}_{\{ h_{\tilde D_N, B_i}  \, \leq \, h_0(B_i) \}} \right] \\
&\leq C_{\mathrm{Prop} \ref{prop:scat_zeros}}(\theta') \; \EE_{\nu_N} \left[ h_{D_N, B_i} \right]^{\tau - \beta}  \;\left( 1 + C_\mathrm{low}(U)^{-1 } C_\mathrm{up}(U) \right) \|f\|_{H^\tau(U)}.
\end{align*}
This finishes the proof of the second claim, with
\[
C_{\mathrm{Thm} \ref{thm:gp_conv_weighted} }' =\left( 1 + C_\mathrm{low}(U)^{-1 } C_\mathrm{up}(U) \right) \|f\|_{H^\tau(U)} \left(1 + r C_{\mathrm{Prop} \ref{prop:scat_zeros}}(\theta') \right).
\]

\end{proof}

\end{document}